\documentclass[10pt,twocolumn,letterpaper]{article}

\usepackage[pagenumbers]{iccv}

\usepackage{amsmath}
\usepackage{amssymb}

\usepackage{times}
\usepackage{graphicx}
\usepackage{multirow}
\usepackage{booktabs}
\usepackage{makecell}
\usepackage{pifont}
\usepackage{marvosym}
\usepackage{fontawesome}
\usepackage{titletoc}

\newcommand{\cmark}{\ding{51}}
\newcommand{\xmark}{\ding{55}}

\definecolor{term_blue}{RGB}{99,113,250}
\definecolor{term_red}{RGB}{239,99,75}
\definecolor{term_green}{RGB}{0,180,139}
\definecolor{term_gray}{RGB}{165,165,165}

\usepackage{soul}
\newcommand{\hlblue}[1]{{\sethlcolor{term_blue!20}\hl{#1}}}
\newcommand{\hlred}[1]{{\sethlcolor{term_red!20}\hl{#1}}}
\newcommand{\hlgreen}[1]{{\sethlcolor{term_green!20}\hl{#1}}}
\newcommand{\hlgray}[1]{{\sethlcolor{term_gray!20}\hl{#1}}}

\definecolor{nu_barrier}{RGB}{112,128,144}
\definecolor{nu_bicycle}{RGB}{220,20,60}
\definecolor{nu_bus}{RGB}{255,0,0}
\definecolor{nu_car}{RGB}{255,158,0}
\definecolor{nu_cons}{RGB}{233,150,70}
\definecolor{nu_motor}{RGB}{255,61,99}
\definecolor{nu_ped}{RGB}{0,0,230}
\definecolor{nu_cone}{RGB}{47,79,79}
\definecolor{nu_trailer}{RGB}{255,140,0}
\definecolor{nu_truck}{RGB}{255,99,71}
\definecolor{nu_driv}{RGB}{0,207,191}
\definecolor{nu_flat}{RGB}{175,0,75}
\definecolor{nu_sidewalk}{RGB}{75,0,75}
\definecolor{nu_terrain}{RGB}{112,180,60}
\definecolor{nu_manmade}{RGB}{222,184,135}
\definecolor{nu_veg}{RGB}{0,175,0}

\definecolor{iccvblue}{rgb}{0.21,0.49,0.74}
\usepackage[pagebackref,breaklinks,colorlinks,allcolors=iccvblue]{hyperref}

\newcommand\blfootnote[1]{%
\begingroup
\renewcommand\thefootnote{}{}\footnote{#1}%
\addtocounter{footnote}{-1}%
\endgroup
}

\def\paperID{127}
\def\confName{ICCV}
\def\confYear{2025}

\title{
\vspace{-0.5cm}
Beyond One Shot, Beyond One Perspective: \\Cross-View and Long-Horizon Distillation for Better LiDAR Representations
}

\author{
    Xiang Xu$^{1}$ \quad Lingdong Kong$^{2,*}$ \quad Song Wang$^{3}$ \quad Chuanwei Zhou$^{4}$ \quad Qingshan Liu$^{4,5,\textrm{\Letter}}$
    \\[0.8ex]
    {\small$^1$Nanjing University of Aeronautics and Astronautics \quad $^2$National University of Singapore \quad $^3$Zhejiang University}
    \\
    {\small$^4$Nanjing University of Posts and Telecommunications \quad $^5$SKL-TI}
    \\[0.8ex]
    \faGithub~\textbf{Code \& Project:} \url{https://github.com/Xiangxu-0103/LiMA}
}

\begin{document}

\maketitle

\blfootnote{$(*)$ Project lead. $(\textrm{\Letter})$ Corresponding author.}

\begin{abstract}
    LiDAR representation learning aims to extract rich structural and semantic information from large-scale, readily available datasets, reducing reliance on costly human annotations. However, existing LiDAR representation strategies often overlook the inherent spatiotemporal cues in LiDAR sequences, limiting their effectiveness. In this work, we propose \textbf{LiMA}, a novel long-term image-to-\underline{\textbf{Li}}DAR \underline{\textbf{M}}emory \underline{\textbf{A}}ggregation framework that explicitly captures longer range temporal correlations to enhance LiDAR representation learning. LiMA comprises three key components: \textbf{1)} a \textbf{Cross-View Aggregation} module that aligns and fuses overlapping regions across neighboring camera views, constructing a more unified and redundancy-free memory bank; \textbf{2)} a \textbf{Long-Term Feature Propagation} mechanism that efficiently aligns and integrates multi-frame image features, reinforcing temporal coherence during LiDAR representation learning; and \textbf{3)} a \textbf{Cross-Sequence Memory Alignment} strategy that enforces consistency across driving sequences, improving generalization to unseen environments. LiMA maintains \textbf{high pretraining efficiency} and incurs no additional computational overhead during downstream tasks.  Extensive experiments on mainstream LiDAR-based perception benchmarks demonstrate that LiMA significantly improves both LiDAR semantic segmentation and 3D object detection. We hope this work inspires more effective pretraining paradigms for autonomous driving. The code has be made publicly accessible for future research.
\end{abstract}

\vspace{-0.5cm}
\section{Introduction}
\label{sec:intro}

LiDAR sensors provide high-resolution spatial information and are essential for precise environmental perception and safe navigation \cite{sun2024lidarseg,kong2025calib3d,bian2025dynamiccity,li2024is}. However, achieving accurate perception relies on large-scale, densely labeled datasets, which are costly and labor-intensive to acquire, thereby limiting scalability in real-world applications \cite{gao2021survey,fei2023survey,guo2020survey,wang2025pointlora}.

\begin{figure}[t]
    \centering
    \includegraphics[width=\linewidth]{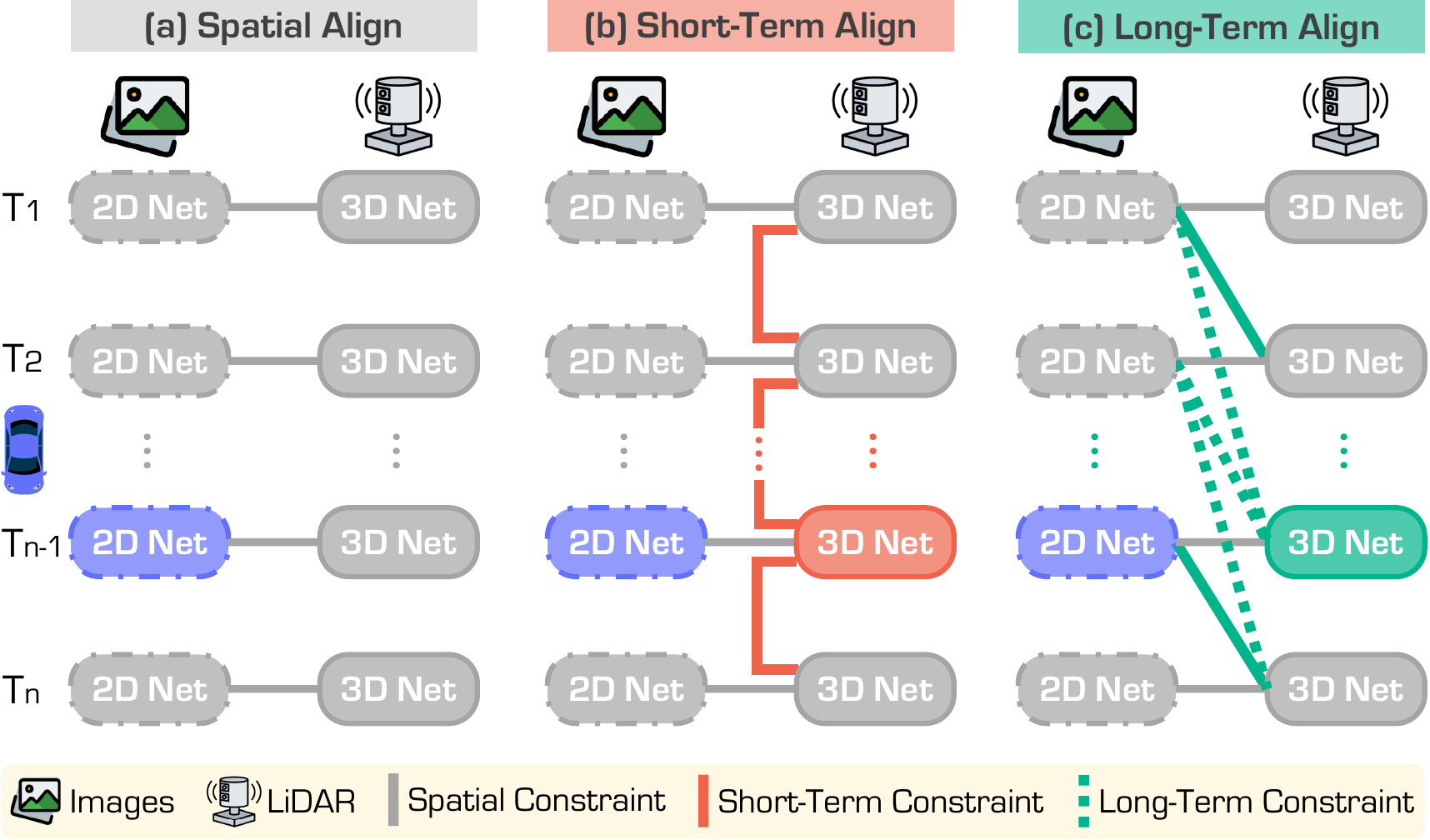}
    \vspace{-0.68cm}
    \caption{\textbf{Illustrative examples of image-to-LiDAR pretraining paradigms}. (a) \hlgray{Spatial Alignment} aligns LiDAR features with corresponding image features in the spatial domain without considering temporal consistency. (b) \hlred{Short-Term} methods propagate LiDAR features frame by frame, ensuring feature consistency across neighboring frames, but fail to capture long-term dependencies. (c) Our approach leverages \hlgreen{Long-Term} image sequences to enrich LiDAR representations, enabling a more comprehensive understanding of long-range dependencies and motion patterns.}
    \label{fig:teaser}
    \vspace{-0.51cm}
\end{figure}

To mitigate this challenge, recent advances in data representation learning leverage large-scale, easily accessible datasets to explore inherent semantic structures \cite{fei2023survey,xiao2023survey,xiao2024survey,liu2024m3net}. Among these approaches, image-to-LiDAR pretraining methods \cite{sautier2022slidr,puy2024scalr,mahmoud2023st-slidr,zhang2024hvdistill} utilize rich visual priors from RGB images to improve LiDAR feature learning. However, as illustrated in \cref{fig:teaser}(a), these methods primarily focus on spatial alignment while overlooking the temporal dynamics in LiDAR sequences. As a result, they struggle to capture motion patterns and scene evolution, which are crucial for robust perception in dynamic environments \cite{hu2022stp3,liu2024lidar,xie2025drivebench,hao2024is}.

Recent works \cite{liu2023seal,pang2023tricc,xu2024superflow} incorporate short-term temporal modeling to mitigate this limitation. As shown in \cref{fig:teaser}(b), these approaches enforce feature consistency across adjacent frames, capturing local motion cues and ensuring smooth feature transitions. However, they rely primarily on frame-to-frame propagation, which maintains local consistency but lacks the capacity to model long-range dependencies and global scene transformations over extended durations. This limitation hinders the ability to construct stable, high-level representations crucial for understanding long-term motion trends and structural variations.

Since LiDAR-based perception tasks inherently involve sequential data, long-term modeling is essential for capturing stable, high-level features that encode extended motion trajectories and scene transitions \cite{wang2023streampetr,lin2022sparse4d,liu2023petrv2}. Unlike short-term methods that focus on immediate feature alignment, long-term modeling provides a more comprehensive understanding of dynamic environments \cite{xie2025robobev}. This is particularly beneficial for tasks such as motion forecasting, behavior prediction, and autonomous decision-making \cite{feng2025survey,tang2022prediction,rudenko2018joint}.

To bridge this gap, we propose long-term image-to-\textbf{Li}DAR \textbf{M}emory \textbf{A}ggregation (\textbf{LiMA}), a simple yet effective pretraining framework that explicitly models temporal structures in driving sequences, as illustrated in \cref{fig:teaser}(c). LiMA introduces three key components to enhance LiDAR representation learning from long-term image sequences.
\begin{itemize}
    \item \textbf{Cross-View Aggregation.} To mitigate redundancy in overlapping camera views, this module identifies and unifies shared regions across multiple viewpoints, reinforcing spatial consistency in LiDAR representations. By aligning and fusing features, it constructs a high-quality memory bank for long-term feature propagation.
    \item \textbf{Long-Term Feature Propagation.} To capture long-range dependencies, stored features in the memory bank are transformed into the ego-vehicle coordinate frame, ensuring feature alignment across frames for seamless temporal fusion. This process establishes a temporally coherent image representation, which is distilled into the LiDAR model to encode motion patterns from extended image sequences. The memory bank is dynamically updated in a first-in, first-out (FIFO) manner, enabling continuous temporal propagation across LiDAR frames.
    \item \textbf{Cross-Sequence Memory Alignment.} To improve robustness across diverse driving conditions, this strategy synthesizes mixed scenes from distinct sequences, facilitating adaptation to varying environments. By maintaining structural coherence with sequence-specific memory banks, it ensures robust feature alignment and improves generalization across heterogeneous driving scenarios.
\end{itemize}
By integrating these three components, LiMA effectively captures both spatial and temporal correlations within and across driving sequences. Extensive experiments on multiple 3D perception benchmarks validate its effectiveness, demonstrating substantial \textbf{performance gains} in both LiDAR semantic segmentation and object detection tasks. Notably, Our framework ensures \textbf{high pretraining efficiency} with an increased number of frames, requiring no more than $20$ hours for pretraining. Furthermore, LiMA introduces no additional computational overhead during downstream tasks, ensuring efficient deployment.

To summarize, this work makes contributions as follows:
\begin{itemize}
    \item We propose \textbf{LiMA}, a long-term image-to-LiDAR memory aggregation framework that captures extended temporal dependencies to improve LiDAR representations.

    \item We introduce three key components: a cross-view aggregation module, a long-term feature propagation module, and a cross-sequence memory alignment strategy, which collectively enhance spatial consistency and temporal robustness across both intra- and inter-sequence domains.

    \item Extensive experiments across multiple 3D perception benchmarks demonstrate the effectiveness of our method, achieving significant performance improvements in LiDAR semantic segmentation and object detection tasks.
\end{itemize}

\section{Related Work}
\label{sec:related_work}

\noindent\textbf{LiDAR Scene Understanding.} LiDAR sensors provide high-precision 3D environmental representations essential for autonomous driving \cite{chen2023clip2Scene,kong2023robodepth,chen2023towards,li2025seeground}. However, the sparsity and irregularity of LiDAR point clouds pose challenges for accurate perception. To address this, existing methods transform point clouds into various representations, including raw points \cite{qi2017pointnet,qi2017pointnet++,hu2020randla,thomas2019kpconv,shuai2021baflac,shi2019pointrcnn}, bird's-eye view (BEV) \cite{zhang2020polarnet,zhou2021panoptic-polarnet,lang2019pointpillars,chen2021polarstream}, range images \cite{milioto2019rangenet++,xu2020squeezesegv3,ando2023rangevit,kong2023rangeformer,xu2025frnet}, sparse voxels \cite{choy2019minkunet,tang2020spvcnn,zhu2021cylinder3d,hong2021dsnet,hong20244ddsnet}, and multi-view formats \cite{xu2021rpvnet,zhuang2021pmf,liu2023uniseg,cheng2021af2s3net,yan20222dpass,xu2024visual,peng2024learning}. While these methods achieve strong performance, they rely heavily on large-scale labeled datasets, which are expensive and time-consuming to obtain. To mitigate this challenge, recent studies have explored semi-supervised \cite{kong2023lasermix,kong2025lasermix++,li2023lim3d,wu2024patchteacher} and weakly-supervised \cite{liu2022box2seg,unal2022scribblekitti,xu2020weakly,meng2021towards,wang2025nuc-net} learning strategies to reduce annotation costs while maintaining high performance in LiDAR-based perception tasks.

\noindent\textbf{Image-to-LiDAR Data Pretraining.} Pretraining facilitates effective feature learning by leveraging tailored objectives such as mask modeling \cite{pang2022masked,yu2022point-bert}, contrastive learning \cite{xie2020pointcontrast,sautier2024bevcontrast}, and reconstruction \cite{huang2023ponder,zheng2024point}. However, early methods are limited to single-modal point clouds, limiting their scalability and adaptability in large-scale driving environments. To address this constraint, SLidR \cite{sautier2022slidr} introduces a pioneering cross-sensor contrastive learning, aligning pretrained image and corresponding LiDAR features. Building on this foundation, subsequent works have incorporated more advanced techniques, including class balance strategies \cite{mahmoud2023st-slidr}, semantic-coherent superpixels \cite{liu2023seal,xu2024superflow,kong2025largead,xu2025superflow++}, hybrid representations \cite{zhang2024hvdistill,xu2025limoe}, and knowledge distillation \cite{puy2024scalr}. Despite these advancements, existing approaches largely overlook the temporal dynamics within LiDAR sequences, which are crucial for capturing motion patterns.

\noindent\textbf{Temporal Modeling for LiDAR Representation.} LiDAR data inherently capture spatiotemporal dynamics, yet early research predominantly focused on object- or human-centric point clouds \cite{chen20224dcontrast,liu2023leaf,huang2021strl,nunes2023tarl,wu2023stssl}, limiting scalability in the highly dynamic and unstructured environments of autonomous driving. To enhance temporal coherence across consecutive scans, recent works have explored various strategies. TriCC \cite{pang2023tricc} enforces triangle consistency to learn temporal relationships, preserving structural integrity over time. Seal \cite{liu2023seal} utilizes RANSAC-based object segmentation \cite{fischler1981ransac} across LiDAR frames to facilitate contrastive learning. SuperFlow \cite{xu2024superflow} enhances temporal modeling by estimating semantic flow across scans, effectively capturing motion cues. However, these methods primarily operate in a frame-by-frame manner, which limits their ability to model long-term dependencies and global scene evolution. In this work, we propose a novel framework that implicitly incorporates long-range temporal information from image sequences into LiDAR representation learning, enabling a more comprehensive understanding of spatiotemporal relationships beyond local pairwise constraints.

\section{Revisit Image-to-LiDAR Data Pretraining}
\label{sec:revisit}

In this section, we revisit common strategies for image-to-LiDAR pretraining, including contrastive learning, knowledge distillation, and temporal modeling.

\subsection{Preliminaries}

\noindent\textbf{Image-to-LiDAR Calibration.} Autonomous driving systems integrate LiDAR and multiple cameras, requiring precise spatial and temporal alignment for effective multimodal perception. Temporal alignment ensures both sensors capture the scene simultaneously, mitigating motion-induced artifacts. Spatial alignment involves estimating extrinsic parameters to transform LiDAR point clouds into the camera frame, enabling accurate feature fusion.

Formally, let $\mathcal{P}^{t} \in \mathbb{R}^{N \times 4}$ denote a point cloud with $N$ points at timestamp $t$, where each point $\mathbf{p}_{i}^{t}$ consists of spatial coordinates $(x_{i}^{t}, y_{i}^{t}, z_{i}^{t})$ and intensity $r_{i}^{t}$. The corresponding camera image $\mathcal{I}^{t} \in \mathbb{R}^{H \times W \times 3}$ has a resolution of $H \times W$. Each LiDAR point $\mathbf{p}_i$ is projected onto the camera plane $(u_{i}^{t}, v_{i}^{t})$ via the following formulation:
\begin{equation}
    \label{eq:calib}
    \left[u_{i}^{t}, v_{i}^{t}\right]^{\text{T}} = \frac{1}{z_{i}^{t}} \times \Gamma_{K}^{t} \times \Gamma_{C}^{t} \times \left[x_{i}^{t}, y_{i}^{t}, z_{i}^{t}\right]^{\text{T}}~,
\end{equation}
where $\Gamma_{K}^{t}$ is the camera intrinsic matrix and $\Gamma_{C}^{t}$ is the extrinsic transformation from LiDAR to the camera frame. This process establishes a set of point-pixel correspondences $\left\{\mathbf{p}_{i}, \mathbf{c}_{i}\right\}_{i=1}^{M}$, where $M$ denotes the number of valid projections within the image bounds.

\noindent\textbf{Pretraining Objective.} The image network $\mathcal{G}_{\theta_{i}}: \mathbb{R}^{H \times W \times 3} \to \mathbb{R}^{H \times W \times C}$, with pretrained parameters $\theta_{i}$, extracts a $C$-dimensional feature map $\mathcal{F}_{i}^{t}$ from $\mathcal{I}^{t}$. Similarly, the LiDAR network $\mathcal{G}_{\theta_{p}}: \mathbb{R}^{N \times 4} \to \mathbb{R}^{N \times C}$, parameterized by $\theta_{p}$, extracts point features $\mathcal{F}_{p}^{t}$ from $\mathcal{P}^{t}$. The pretraining objective optimizes $\theta_{p}$ such that $\mathcal{F}_{p}^{t}$ aligns with $\mathcal{F}_{i}^{t}$, allowing the LiDAR model to inherit semantic priors from the image domain while reducing reliance on labeled datasets.

\subsection{Pretraining Methods}

\noindent\textbf{Contrastive Learning.} Contrastive learning improves cross-modal alignment by constructing positive and negative feature pairs. In image-to-LiDAR learning, contrastive losses can be formulated at the point-pixel \cite{liu2021ppkt} or superpoint-superpixel \cite{sautier2022slidr,mahmoud2023st-slidr,liu2023seal,xu2024superflow} levels:
\begin{equation}
    \label{eq:contrastive_loss}
    \mathcal{L}_{\mathrm{cont}}(\mathcal{F}_{i}^{t}, \mathcal{F}_{p}^{t}) = \frac{1}{M} \sum_{j=1}^{M} \log \frac{e^{\langle \mathbf{f}_{(i,j)}^{t}, \mathbf{f}_{(p,j)}^{t} \rangle / \tau}}{\sum_{k=1}^{M} e^{\langle \mathbf{f}_{(i,k)}^{t}, \mathbf{f}_{(p,j)}^{t} \rangle / \tau}}~,
\end{equation}
where $\langle \cdot, \cdot \rangle$ denotes the dot product of feature embeddings, and $\tau > 0$ is a temperature scaling factor. This loss encourages positive feature pairs to be closer while pushing apart negatives, fostering modality-invariant representations.

\noindent\textbf{Knowledge Distillation.} Knowledge distillation transfers knowledge from a pretrained image model to a LiDAR model, enforcing feature consistency \cite{puy2024scalr}. A common objective is the $\ell_2$ loss, directly operating in the feature space:
\begin{equation}
    \label{eq:distillation_loss}
    \mathcal{L}_{\mathrm{dist}}(\mathcal{F}_{i}^{t}, \mathcal{F}_{p}^{t}) = \frac{1}{M} \sum\nolimits_{j=1}^{M} \|\mathbf{f}_{(i,j)}^{t} - \mathbf{f}_{(p,j)}^{t}\|_{2}~.
\end{equation}
By minimizing feature discrepancies, this loss aligns image and LiDAR features, allowing the LiDAR model to leverage image priors while maintaining computational efficiency.

\noindent\textbf{Temporal Modeling.} Temporal modeling captures dependencies across consecutive LiDAR scans. To enforce feature consistency across frames, objects are first segmented, and temporal contrastive learning is applied \cite{liu2023seal,xu2024superflow}:
\begin{equation}
    \label{eq:temporal_loss}
    \mathcal{L}_{\mathrm{cont}}(\mathcal{F}_{p}^{t}, \mathcal{F}_{p}^{t+1}) = \frac{1}{S} \sum_{j=1}^{S} \log \frac{e^{\langle \mathbf{f}_{(p,j)}^{t}, \mathbf{f}_{(p,j)}^{t+1} \rangle / \tau}}{\sum_{k=1}^{S} e^{\langle \mathbf{f}_{(p,k)}^{t}, \mathbf{f}_{(p,j)}^{t+1} \rangle / \tau}}~,
\end{equation}
where $S$ is the number of segmented objects. By maximizing feature similarity between corresponding objects across frames while contrasting them with non-corresponding ones, this loss enforces temporal coherence, enhancing motion-awareness in LiDAR representations.

\begin{figure*}[t]
    \centering
    \includegraphics[width=1.0\linewidth]{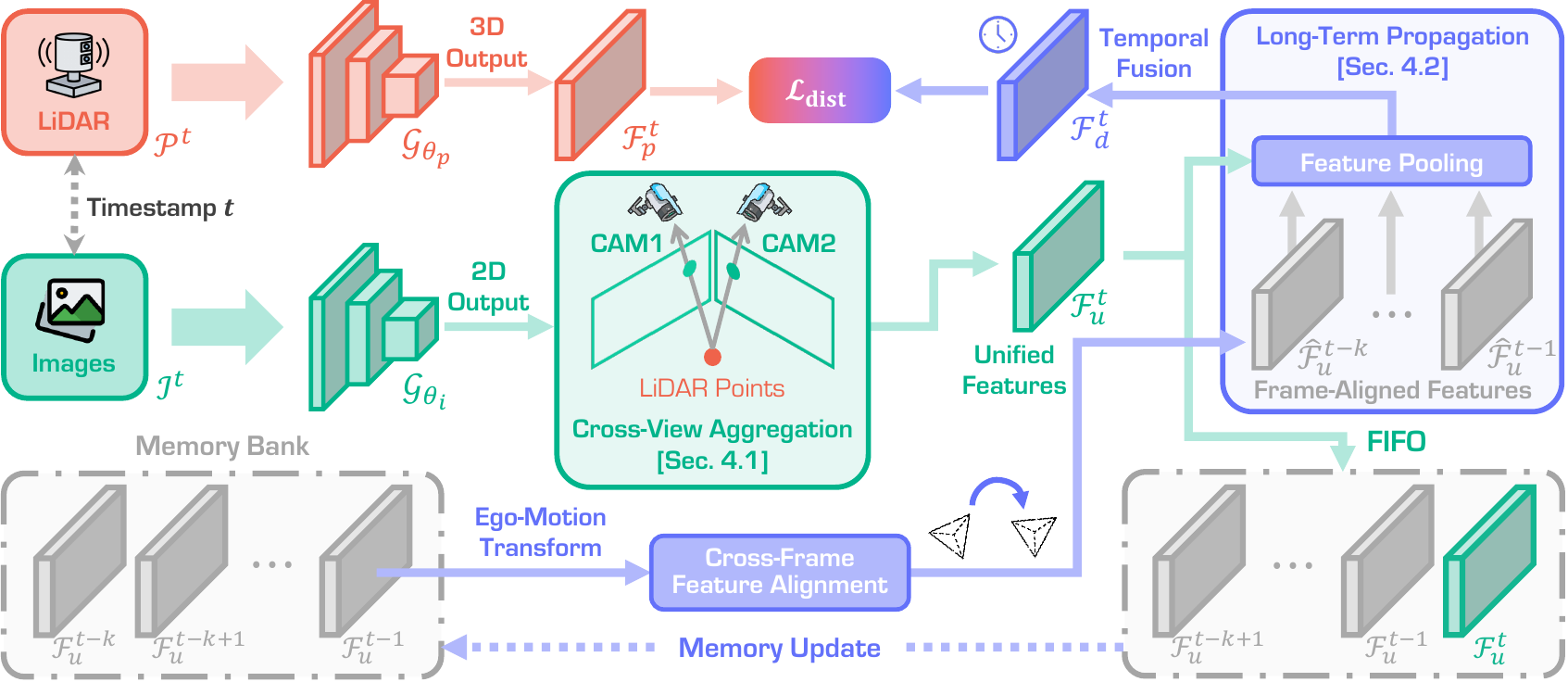}
    \vspace{-0.7cm}
    \caption{\textbf{Overview of the LiMA framework}. At each timestamp $t$, multi-view image features are first extracted and unified through the \hlgreen{Cross-View Aggregation} module, producing $\mathcal{F}_{u}^{t}$ (\cref{sec:cva}). A \hlgray{Memory Bank} is introduced to maintain unified image features from the past $k$ frames, enabling temporal feature propagation and fusion with $\mathcal{F}_{u}^{t}$ to capture \hlblue{Long-Term Motion Patterns} $\mathcal{F}_{d}^{t}$ (\cref{sec:term}). The enriched temporal features are distilled into LiDAR representation $\mathcal{F}_{p}^{t}$, enabling \hlred{Cross-Modal Learning}. The memory bank is continuously updated in a first-in, first-out (FIFO) manner, ensuring effective feature propagation and refinement for future frames.}
    \label{fig:framework}
    \vspace{-0.4cm}
\end{figure*}

\section{Methodology}
\label{sec:method}

As revisited in \cref{sec:revisit}, existing approaches \cite{sautier2022slidr,puy2024scalr,liu2023seal,xu2024superflow} focus on spatial alignment or short-term feature propagation, neglecting long-term dependencies, which are crucial for modeling stable representations and extended scene transitions. To address these limitations, we introduce long-term image-to-\textbf{Li}DAR \textbf{M}emory \textbf{A}ggregation (\textbf{LiMA}), a simple yet effective framework that distills rich motion patterns from longer sequences into LiDAR representations, enhancing the robustness and generalizability in LiDAR data pretraining. The overall architecture of LiMA is illustrated in \cref{fig:framework}. It comprises three key components: \textbf{cross-view aggregation}, which unifies multi-view image features to construct a high-quality memory bank (\cf \cref{sec:cva}), a \textbf{long-term feature propagation} module, which enforces temporal consistency and propagates long-term motion cues (\cf \cref{sec:term}), and a \textbf{cross-sequence memory alignment} mechanism, which improves robustness by aligning long-term features across diverse driving contexts (\cf \cref{sec:cross-sequence}).

\subsection{Cross-View Aggregation}
\label{sec:cva}

\noindent\textbf{Motivation.} Autonomous vehicles utilize multiple cameras to achieve a surround-view perception of their environment, ensuring comprehensive scene coverage. However, due to the inherent overlap in neighboring cameras' fields of view, redundant yet complementary visual information is captured across multiple perspectives. When LiDAR points are projected onto these camera views, they may be mapped to multiple pixel locations, resulting in inconsistent feature representations. This multi-view ambiguity introduces conflicts during feature alignment, leading to optimization instability and degraded learning efficiency.

\noindent\textbf{Cross-View Aggregation.} To address this challenge, we propose a cross-view aggregation mechanism that unifies feature representations across different camera perspectives. Specifically, for each LiDAR point appearing in multiple views, we extract the corresponding pixel-aligned features $\{\mathbf{f}^{t}_{(i,j)}\}_{j=1}^{V}$ from $V$ cameras and aggregate them with an averaging operation to obtain the unified feature representation $\mathcal{F}^{t}_{u}$. This operation produces a unified feature representation that mitigates inconsistencies while preserving complementary information from different perspectives. Compared to alternative fusion strategies, such as max or attention-based aggregation, our mean aggregation provides a balanced fusion of cross-view information while naturally adapting to variations in sensor configurations, ensuring stable and efficient multi-view feature integration.

\noindent\textit{Role in Our Framework.} The cross-view aggregation module enhances the spatial coherence of LiDAR features by unifying multi-view representations, mitigating optimization conflicts, and ensuring stable and efficient training. This strengthens the model’s capacity to learn robust and invariant representations, providing a more consistent and informative feature space for the memory bank.

\subsection{Long-Term Feature Propagation}
\label{sec:term}

\noindent\textbf{Motivation.} Capturing long-range temporal dependencies is crucial for robust autonomous driving perception \cite{lin2022sparse4d,wang2023streampetr,liu2023petrv2}. However, directly extracting features from historical frames at every timestep incurs significant computational and memory overhead. To tackle this challenge, we propose a long-term feature propagation strategy that efficiently retains and aggregates informative representations from past frames. By ensuring temporal alignment and reducing redundant computations, this approach enables LiDAR features to capture rich motion-aware contextual information while maintaining computational efficiency.

\noindent\textbf{Long-Term Feature Alignment \& Propagation.} At each timestamp $t$, we maintain a structured memory bank that stores compact yet informative unified image features from the past $k$ frames, denoted as $\{\mathcal{F}_{u}^{t-k},\dots,\mathcal{F}_{u}^{t-1}\}$. To ensure temporal alignment, we first apply ego-motion transformation, warping historical features into the current frame's coordinate space through temporal calibration \cite{wang2023streampetr}. This yields a set of frame-aligned features $\{\widehat{\mathcal{F}}_{u}^{t-k},\dots,\widehat{\mathcal{F}}_{u}^{t-1}\}$, effectively compensating for vehicle motion and viewpoint variations. The aligned features are then efficiently retrieved and aggregated with the current frame's features $\mathcal{F}_{u}^{t}$ via average pooling, resulting in a temporally enriched representation $\mathcal{F}_{d}^{t}$. Finally, $\mathcal{F}_{d}^{t}$ is distilled into LiDAR representations $\mathcal{F}_{p}^{t}$ to facilitate the pretraining objective, as depicted in \cref{eq:distillation_loss}. To ensure computational efficiency, the memory bank employs a first-in, first-out (FIFO) update strategy, continuously storing the latest $\mathcal{F}_{u}^{t}$ while discarding obsolete frames, thereby preserving essential long-term information without incurring excessive memory overhead.

\noindent\textit{Role in Our Framework.} Long-term feature propagation serves as a cornerstone of LiMA, enabling the model to capture temporal dynamics efficiently while avoiding redundant computations. By integrating motion-aware contexture information across frames, our approach significantly enhances the spatial-temporal consistency of LiDAR feature learning. Unlike conventional methods that process each frame independently, LiMA exploits historical context to improve generalization within driving sequences.

\subsection{Cross-Sequence Memory Alignment}
\label{sec:cross-sequence}

\begin{figure}[t]
    \centering
    \includegraphics[width=\linewidth]{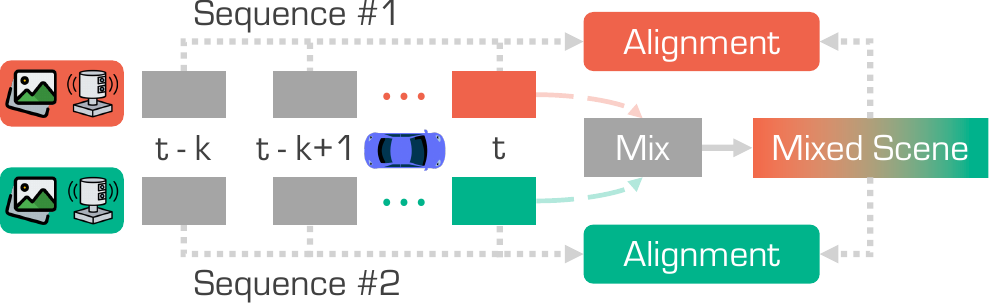}
    \vspace{-0.75cm}
    \caption{Illustration of \textbf{cross-sequence memory alignment}. Existing mixed strategies \cite{kong2023lasermix,xiao2022polarmix} are utilized to blend two LiDAR scenes from distinct sequences, generating out-of-distribution variations. This process encourages feature alignment across sequences by leveraging memory banks from the original sequences.}
    \label{fig:cross_alignment}
    \vspace{-0.45cm}
\end{figure}

\noindent\textbf{Motivation.} While our framework effectively retains historical context within a single driving sequence, its reliance on intra-sequence consistency constrains its ability to generalize across diverse real-world scenarios. Autonomous driving environments undergo substantial variations in weather, illumination, and road structures, necessitating a mechanism that enforces feature alignment not only within a sequence but also across different driving contexts.

\noindent\textbf{Cross-Sequence Adaptation.} To address this challenge, we propose a cross-sequence memory alignment strategy designed to enhance feature consistency and improve generalization across diverse driving scenarios. As illustrated in \cref{fig:cross_alignment}, given two LiDAR point clouds $\mathcal{P}_{1}$ and $\mathcal{P}_{2}$ from distinct sequences, we leverage established mixing strategies, such as LaserMix \cite{kong2023lasermix} and PolarMix \cite{xiao2022polarmix}, to generate a synthetic mixed scene $\mathcal{P}_{m}$. This mixed representation introduces controlled out-of-distribution variations, facilitating alignment across heterogeneous driving conditions. To reinforce feature consistency, the mixed features are optimized to maintain structural coherence with the original sequence memory banks, ensuring robust adaptation across both spatial and temporal domains.

\noindent\textit{Role in Our Framework.} 
The proposed cross-sequence memory alignment functions as a mixed-training strategy designed to bridge domain gaps across driving sequences, maintain long-term feature coherence, and enhance generalization in dynamic real-world settings. By integrating cross-sequence consistency into the learning process, LiMA significantly improves the robustness and transferability of LiDAR-based representations, enabling more reliable downstream perception across diverse environments.

\begin{table*}[t]
    \centering
    \caption{\textbf{Comparisons of state-of-the-art LiDAR pretraining methods} pretrained on \textit{nuScenes} \cite{caesar2020nuscenes} and fine-tuned on \textit{nuScenes} \cite{fong2022panoptic-nuscenes}, \textit{SemanticKITTI} \cite{behley2019semantickitti}, and \textit{Waymo Open} \cite{sun2020waymo} datasets, respectively, with specific data portions. \textbf{LP} denotes linear probing with frozen backbones. All scores are given in percentage (\%). The \hlgreen{Best} and \hlred{2nd Best} scores under each group are highlighted in \hlgreen{Green} and \hlred{Red}.}
    \vspace{-0.3cm}
    \label{tab:benchmark}
    \resizebox{\linewidth}{!}{
    \begin{tabular}{r|r|c|c|c|p{30pt}<{\centering}p{30pt}<{\centering}p{30pt}<{\centering}p{30pt}<{\centering}p{30pt}<{\centering}p{30pt}<{\centering}|p{36pt}<{\centering}|p{36pt}<{\centering}}
        \toprule
        \multirow{2}{*}{\textbf{Method}} & \multirow{2}{*}{\textbf{Venue}} & \textbf{Backbone} & \textbf{Backbone} & \multirow{2}{*}{\textbf{Frames}} & \multicolumn{6}{c|}{\textbf{nuScenes}} & \textbf{KITTI} & \textbf{Waymo}
        \\
        & & \textbf{(2D)} & \textbf{(3D)} & & \textbf{LP} & \textbf{1\%} & \textbf{5\%} & \textbf{10\%} & \textbf{25\%} & \textbf{Full} & \textbf{1\%} & \textbf{1\%}
        \\\midrule\midrule
        \textcolor{gray}{Random} & \textcolor{gray}{-} & \textcolor{gray}{-} & \textcolor{gray}{-} & \textcolor{gray}{-} & \textcolor{gray}{${8.10}$} & \textcolor{gray}{${30.30}$} & \textcolor{gray}{${47.84}$} & \textcolor{gray}{${56.15}$} & \textcolor{gray}{${65.48}$} & \textcolor{gray}{${74.66}$} & \textcolor{gray}{${39.50}$} & \textcolor{gray}{${39.41}$}
        \\\midrule
        SLidR \cite{sautier2022slidr} & CVPR'22 & \multirow{6}{*}{\makecell{ResNet-50 \\ \cite{he2016resnet}}} & \multirow{6}{*}{\makecell{MinkUNet-34 \\ \cite{choy2019minkunet}}} & $1$ & $38.80$ & $38.30$ & $52.49$ & $59.84$ & $66.91$ & $74.79$ & $44.60$ & \hlred{$\mathbf{47.12}$}
        \\
        ST-SLidR \cite{mahmoud2023st-slidr} & CVPR'23 & & & $1$ & $40.48$ & $40.75$ & $54.69$ & $60.75$ & $67.70$ & $75.14$ & $44.72$ & $44.93$
        \\
        TriCC \cite{pang2023tricc} & CVPR'23 & & & $2$ & $38.00$ & $41.20$ & $54.10$ & $60.40$ & $67.60$ & $75.60$ & $45.90$ & -
        \\
        Seal \cite{liu2023seal} & NeurIPS'23 & & & $2$ & \hlred{$\mathbf{44.95}$} & \hlred{$\mathbf{45.84}$} & $55.64$ & \hlred{$\mathbf{62.97}$} & $68.41$ & $75.60$ & $46.63$ & \hlgreen{$\mathbf{49.34}$}
        \\
        CSC \cite{chen2024csc} & CVPR'24 & & & $1$ & \hlgreen{$\mathbf{46.00}$} & \hlgreen{$\mathbf{47.00}$} & \hlgreen{$\mathbf{57.00}$} & \hlgreen{$\mathbf{63.30}$} & \hlred{$\mathbf{68.60}$} & \hlred{$\mathbf{75.70}$} & \hlred{$\mathbf{47.20}$} & -
        \\
        HVDistill \cite{zhang2024hvdistill} & IJCV'24 & & & $1$ & $39.50$ & $42.70$ & \hlred{$\mathbf{56.60}$} & $62.90$ & \hlgreen{$\mathbf{69.30}$} & \hlgreen{$\mathbf{76.60}$} & \hlgreen{$\mathbf{49.70}$} & -
        \\\midrule
        SLidR \cite{sautier2022slidr} & CVPR'22 & \multirow{5}{*}{\makecell{ViT-S \\ \cite{dosovitskiy2021vit}}} & \multirow{5}{*}{\makecell{MinkUNet-34 \\ \cite{choy2019minkunet}}} & $1$ & $44.70$ & $41.16$ & $53.65$ & $61.47$ & $66.71$ & $74.20$ & $44.67$ & $47.57$
        \\
        Seal \cite{liu2023seal} & NeurIPS'23 & & & $2$ & $45.16$ & $44.27$ & $55.13$ & $62.46$ & $67.64$ & $75.58$ & $46.51$ & $48.67$
        \\
        SuperFlow \cite{xu2024superflow} & ECCV'24 & & & $3$ & $46.44$ & \hlred{$\mathbf{47.81}$} & \hlred{$\mathbf{59.44}$} & \hlred{$\mathbf{64.47}$} & \hlred{$\mathbf{69.20}$} & \hlred{$\mathbf{76.54}$} & \hlred{$\mathbf{47.97}$} & \hlred{$\mathbf{49.94}$}
        \\
        ScaLR \cite{puy2024scalr} & CVPR'24 & & & $1$ & \hlred{$\mathbf{49.66}$} & $45.89$ & $56.52$ & $61.07$ & $65.79$ & $73.39$ & $46.06$ & $47.67$
        \\
        \textbf{LiMA} & \textbf{Ours} & & & $6$ & \hlgreen{$\mathbf{54.76}$} & \hlgreen{$\mathbf{48.75}$} & \hlgreen{$\mathbf{60.83}$} & \hlgreen{$\mathbf{65.41}$} & \hlgreen{$\mathbf{69.31}$} & \hlgreen{$\mathbf{76.94}$} & \hlgreen{$\mathbf{49.28}$} & \hlgreen{$\mathbf{50.23}$}
        \\\midrule
        SLidR \cite{sautier2022slidr} & CVPR'22 & \multirow{5}{*}{\makecell{ViT-B \\ \cite{dosovitskiy2021vit}}} & \multirow{5}{*}{\makecell{MinkUNet-34 \\ \cite{choy2019minkunet}}} & $1$ & $45.35$ & $41.64$ & $55.83$ & $62.68$ & $67.61$ & $74.98$ & $45.50$ & $48.32$
        \\
        Seal \cite{liu2023seal} & NeurIPS'23 & & & $2$ & $46.59$ & $45.98$ & $57.15$ & $62.79$ & $68.18$ & $75.41$ & $47.24$ & $48.91$
        \\
        SuperFlow \cite{xu2024superflow} & ECCV'24 & & & $3$ & $47.66$ & $48.09$ & \hlred{$\mathbf{59.66}$} & \hlred{$\mathbf{64.52}$} & \hlred{$\mathbf{69.79}$} & \hlred{$\mathbf{76.57}$} & \hlred{$\mathbf{48.40}$} & \hlred{$\mathbf{50.20}$}
        \\
        ScaLR \cite{puy2024scalr} & CVPR'24 & & & $1$ & \hlred{$\mathbf{51.90}$} & \hlred{$\mathbf{48.90}$} & $57.69$ & $62.88$ & $66.85$ & $74.15$ & $47.77$ & $49.38$
        \\
        \textbf{LiMA} & \textbf{Ours} & & & $6$ & \hlgreen{$\mathbf{56.65}$} & \hlgreen{$\mathbf{51.29}$} & \hlgreen{$\mathbf{61.11}$} & \hlgreen{$\mathbf{65.62}$} & \hlgreen{$\mathbf{70.43}$} & \hlgreen{$\mathbf{76.91}$} & \hlgreen{$\mathbf{50.44}$} & \hlgreen{$\mathbf{51.35}$}
        \\\midrule
        SLidR \cite{sautier2022slidr} & CVPR'22 & \multirow{5}{*}{\makecell{ViT-L \\ \cite{dosovitskiy2021vit}}} & \multirow{5}{*}{\makecell{MinkUNet-34 \\ \cite{choy2019minkunet}}} & $1$ & $45.70$ & $42.77$ & $57.45$ & $63.20$ & $68.13$ & $75.51$ & $47.01$ & $48.60$
        \\
        Seal \cite{liu2023seal} & NeurIPS'23 & & & $2$ & $46.81$ & $46.27$ & $58.14$ & $63.27$ & $68.67$ & $75.66$ & $47.55$ & $50.02$
        \\
        SuperFlow \cite{xu2024superflow} & ECCV'24 & & & $3$ & $48.01$ & \hlred{$\mathbf{49.95}$} & \hlred{$\mathbf{60.72}$} & \hlred{$\mathbf{65.09}$} & \hlred{$\mathbf{70.01}$} & \hlred{$\mathbf{77.19}$} & \hlred{$\mathbf{49.07}$} & \hlred{$\mathbf{50.67}$}
        \\
        ScaLR \cite{puy2024scalr} & CVPR'24 & & & $1$ & \hlred{$\mathbf{51.77}$} & $49.13$ & $58.36$ & $62.75$ & $66.80$ & $74.16$ & $48.64$ & $49.72$
        \\
        \textbf{LiMA} & \textbf{Ours} & & & $6$ & \hlgreen{$\mathbf{56.67}$} & \hlgreen{$\mathbf{53.22}$} & \hlgreen{$\mathbf{62.46}$} & \hlgreen{$\mathbf{66.00}$} & \hlgreen{$\mathbf{70.59}$} & \hlgreen{$\mathbf{77.23}$} & \hlgreen{$\mathbf{52.29}$} & \hlgreen{$\mathbf{51.19}$}
        \\\bottomrule
    \end{tabular}}
    \vspace{-0.2cm}
\end{table*}

\begin{table*}[t!]
    \centering
    \caption{\textbf{Domain generalization study of different LiDAR pretraining methods} pretrained on the \textit{nuScenes} \cite{caesar2020nuscenes} dataset and fine-tuned on a collection of seven different LiDAR semantic segmentation datasets  \cite{unal2022scribblekitti,jiang2021rellis-3d,pan2020semanticposs,xiao2023semanticstf,xiao2022synlidar,klokov2023daps3d,saltori2022synth4d}, respectively, with specific data portions. All scores are given in percentage (\%). The \hlgreen{Best} and \hlred{2nd Best} scores from each metric are highlighted in \hlgreen{Green} and \hlred{Red}.}
    \vspace{-0.3cm}
    \label{tab:multiple_dataset}
    \resizebox{\linewidth}{!}{
    \begin{tabular}{r|r|cc|cc|cc|cc|cc|cc|cc}
        \toprule
        \multirow{2}{*}{\textbf{Method}} & \multirow{2}{*}{\textbf{Venue}} & \multicolumn{2}{c|}{\textbf{ScriKITTI}} & \multicolumn{2}{c|}{\textbf{Rellis-3D}} & \multicolumn{2}{c|}{\textbf{SemPOSS}} & \multicolumn{2}{c|}{\textbf{SemSTF}} & \multicolumn{2}{c|}{\textbf{SynLiDAR}} & \multicolumn{2}{c|}{\textbf{DAPS-3D}} & \multicolumn{2}{c}{\textbf{Synth4D}}
        \\
        & & \textbf{1\%} & \textbf{10\%} & \textbf{1\%} & \textbf{10\%} & \textbf{Half} & \textbf{Full} & \textbf{Half} & \textbf{Full} & \textbf{1\%} & \textbf{10\%} & \textbf{Half} & \textbf{Full} & \textbf{1\%} & \textbf{10\%}
        \\\midrule\midrule
        \textcolor{gray}{Random} & \textcolor{gray}{-} & \textcolor{gray}{${23.81}$} & \textcolor{gray}{${47.60}$} & \textcolor{gray}{${38.46}$} & \textcolor{gray}{${53.60}$} & \textcolor{gray}{${46.26}$} & \textcolor{gray}{${54.12}$} & \textcolor{gray}{${48.03}$} & \textcolor{gray}{${48.15}$} & \textcolor{gray}{${19.89}$} & \textcolor{gray}{${44.74}$} & \textcolor{gray}{${74.32}$} & \textcolor{gray}{${79.38}$} & \textcolor{gray}{${20.22}$} & \textcolor{gray}{${66.87}$}
        \\\midrule
        SLidR \cite{sautier2022slidr} & CVPR'22 & $39.60$ & $50.45$ & $49.75$ & $54.57$ & $51.56$ & $55.36$ & $52.01$ & $54.35$ & $42.05$ & $47.84$ & $81.00$ & $85.40$ & $63.10$ & $62.67$
        \\
        Seal \cite{liu2023seal} & NeurIPS'23 & $40.64$ & $52.77$ & $51.09$ & $55.03$ & $53.26$ & $56.89$ & $53.46$ & $55.36$ & $43.58$ & $49.26$ & $81.88$ & $85.90$ & $64.50$ & $66.96$
        \\
        SuperFlow \cite{xu2024superflow} & ECCV'24 & \hlred{$\mathbf{42.70}$} & \hlred{$\mathbf{54.00}$} & \hlred{$\mathbf{52.83}$} & \hlred{$\mathbf{55.71}$} & \hlred{$\mathbf{54.41}$} & \hlred{$\mathbf{57.33}$} & \hlred{$\mathbf{54.72}$} & \hlred{$\mathbf{56.57}$} & \hlred{$\mathbf{44.85}$} & $51.38$ & \hlred{$\mathbf{82.43}$} & \hlred{$\mathbf{86.21}$} & \hlred{$\mathbf{65.31}$} & \hlred{$\mathbf{69.43}$}
        \\
        ScaLR \cite{puy2024scalr} & CVPR'24 & $40.64$ & $52.39$ & $52.53$ & $55.57$ & $53.65$ & $56.86$ & $54.06$ & $55.96$ & $44.42$ & \hlred{$\mathbf{51.96}$} & $81.92$ & $85.58$ & $64.36$ & $67.44$
        \\
        \textbf{LiMA} & \textbf{Ours} & \hlgreen{$\mathbf{45.90}$} & \hlgreen{$\mathbf{55.13}$} & \hlgreen{$\mathbf{55.62}$} & \hlgreen{$\mathbf{57.15}$} & \hlgreen{$\mathbf{55.05}$} & \hlgreen{$\mathbf{57.81}$} & \hlgreen{$\mathbf{55.45}$} & \hlgreen{$\mathbf{56.70}$} & \hlgreen{$\mathbf{46.66}$} & \hlgreen{$\mathbf{52.32}$} & \hlgreen{$\mathbf{83.11}$} & \hlgreen{$\mathbf{86.63}$} & \hlgreen{$\mathbf{66.04}$} & \hlgreen{$\mathbf{70.19}$}
        \\\bottomrule
    \end{tabular}}
    \vspace{-0.45cm}
\end{table*}

\begin{table*}[t]
    \centering
    \caption{\textbf{Out-of-distribution robustness assessment} of LiDAR pretraining methods under corruptions and sensor failures in the \textit{nuScenes-C} dataset from the \textit{Robo3D} benchmark \cite{kong2023robo3d}. \textbf{Full} denotes fine-tuning with full labels. \textbf{LP} denotes linear probing with frozen backbones. All mCE, mRR, and mIoU scores are given in percentage (\%). The \hlgreen{Best} and \hlred{2nd Best} scores are highlighted in \hlgreen{Green} and \hlred{Red}.}
    \vspace{-0.3cm}
    \label{tab:robo3d}
    \resizebox{\linewidth}{!}{
    \begin{tabular}{c|r|r|p{36pt}<{\centering}|p{36pt}<{\centering}|p{32pt}<{\centering}p{32pt}<{\centering}p{32pt}<{\centering}p{32pt}<{\centering}p{32pt}<{\centering}p{32pt}<{\centering}p{32pt}<{\centering}c|c}
        \toprule
        \textbf{\#} & \textbf{Method} & \textbf{Venue} & \textbf{mCE} $\downarrow$ & \textbf{mRR} $\uparrow$ & \textbf{Fog} $\uparrow$ & \textbf{Rain} $\uparrow$ & \textbf{Snow} $\uparrow$ & \textbf{Blur}$\uparrow$ & \textbf{Beam} $\uparrow$ & \textbf{Cross} $\uparrow$ & \textbf{Echo} $\uparrow$ & \textbf{Sensor} $\uparrow$ & \textbf{Average} $\uparrow$
        \\\midrule\midrule
        \textbf{\textcolor{gray}{Full}} & \textcolor{gray}{Random} & \textcolor{gray}{-} & \textcolor{gray}{${112.20}$} & \textcolor{gray}{${72.57}$} & \textcolor{gray}{${62.96}$} & \textcolor{gray}{${70.65}$} & \textcolor{gray}{${55.48}$} & \textcolor{gray}{${51.71}$} & \textcolor{gray}{${62.01}$} & \textcolor{gray}{${31.56}$} & \textcolor{gray}{${59.64}$} & \textcolor{gray}{${39.41}$} & \textcolor{gray}{${54.18}$}
        \\\midrule
        \multirow{5}{*}{\textbf{Full}} & SLidR \cite{sautier2022slidr} & CVPR'22 & $106.08$ & $75.99$ & $65.41$ & $72.31$ & $56.01$ & $56.07$ & $62.87$ & $41.94$ & $61.16$ & $38.90$ & $56.83$
        \\
        & Seal \cite{liu2023seal} & NeurIPS'23 & $92.63$ & \hlred{$\mathbf{83.08}$} & \hlgreen{$\mathbf{72.66}$} & \hlred{$\mathbf{74.31}$} & $66.22$ & \hlred{$\mathbf{66.14}$} & $65.96$ & \hlred{$\mathbf{57.44}$} & \hlred{$\mathbf{59.87}$} & $39.85$ & $62.81$
        \\
        & SuperFlow \cite{xu2024superflow} & ECCV'24 & \hlred{$\mathbf{91.67}$} & \hlgreen{$\mathbf{83.17}$} & $70.32$ & \hlgreen{$\mathbf{75.77}$} & $65.41$ & $61.05$ & \hlgreen{$\mathbf{68.09}$} & \hlgreen{$\mathbf{60.02}$} & $58.36$ & \hlgreen{$\mathbf{50.41}$} & \hlgreen{$\mathbf{63.68}$}
        \\
        & ScaLR \cite{puy2024scalr} & CVPR'24 & $99.36$ & $80.67$ & $68.43$ & $72.15$ & \hlred{$\mathbf{66.99}$} & $60.07$ & \hlred{$\mathbf{67.35}$} & $44.02$ & $58.98$ & $40.53$ & $59.82$
        \\
        & \textbf{LiMA} & \textbf{Ours} & \hlgreen{$\mathbf{91.43}$} & $82.57$ & \hlred{$\mathbf{71.24}$} & $73.38$ & \hlgreen{$\mathbf{67.33}$} & \hlgreen{$\mathbf{66.73}$} & $66.71$ & $47.66$ & \hlgreen{$\mathbf{61.72}$} & \hlred{$\mathbf{48.65}$} & \hlred{$\mathbf{62.93}$}
        \\\midrule
        \multirow{5}{*}{\textbf{LP}} & SLidR \cite{sautier2022slidr} & CVPR'22 & $179.38$ & $77.18$ & $34.88$ & $38.09$ & $32.64$ & $26.44$ & $33.73$ & $20.81$ & $31.54$ & $21.44$ & $29.95$
        \\
        & Seal \cite{liu2023seal} & NeurIPS'23 & $166.18$ & $75.38$ & $37.33$ & $42.77$ & $29.93$ & $37.73$ & $40.32$ & $20.31$ & $37.73$ & $24.94$ & $33.88$
        \\
        & SuperFlow \cite{xu2024superflow} & ECCV'24 & $161.78$ & $75.52$ & $37.59$ & $43.42$ & $37.60$ & $39.57$ & $41.40$ & $23.64$ & $38.03$ & $26.69$ & $35.99$
        \\
        & ScaLR \cite{puy2024scalr} & CVPR'24 & \hlred{$\mathbf{150.45}$} & \hlred{$\mathbf{78.24}$} & \hlred{$\mathbf{45.27}$} & \hlred{$\mathbf{50.42}$} & \hlred{$\mathbf{41.75}$} & \hlred{$\mathbf{40.69}$} & \hlred{$\mathbf{44.53}$} & \hlgreen{$\mathbf{29.93}$} & \hlred{$\mathbf{41.03}$} & \hlred{$\mathbf{31.22}$} & \hlred{$\mathbf{40.61}$}
        \\
        & \textbf{LiMA} & \textbf{Ours} & \hlgreen{$\mathbf{137.23}$} & \hlgreen{$\mathbf{79.30}$} & \hlgreen{$\mathbf{51.52}$} & \hlgreen{$\mathbf{54.90}$} & \hlgreen{$\mathbf{45.63}$} & \hlgreen{$\mathbf{50.55}$} & \hlgreen{$\mathbf{49.67}$} & \hlred{$\mathbf{27.24}$} & \hlgreen{$\mathbf{45.76}$} & \hlgreen{$\mathbf{34.09}$} & \hlgreen{$\mathbf{44.92}$}
        \\\bottomrule
    \end{tabular}}
    \vspace{-0.4cm}
\end{table*}

\begin{table}[t]
    \centering
    \caption{\textbf{Comparisons of state-of-the-art LiDAR pretraining methods} pretrained and fine-tuned on the \textit{nuScenes} \cite{caesar2020nuscenes} dataset with specific data portions. All detection methods employ CenterPoint \cite{yin2021centerpoint} or SECOND \cite{yan2018second} as the 3D object detection method.}
    \vspace{-0.3cm}
    \label{tab:detection}
    \resizebox{\linewidth}{!}{
    \begin{tabular}{r|r|cc|cc|cc}
        \toprule
        \multirow{3.5}{*}{\textbf{Method}} & \multirow{3.5}{*}{\textbf{Venue}} & \multicolumn{6}{c}{\textbf{nuScenes}}
        \\\cmidrule{3-8}
        & & \multicolumn{2}{c|}{\textbf{5\%}} & \multicolumn{2}{c|}{\textbf{10\%}}& \multicolumn{2}{c}{\textbf{20\%}}
        \\
        & & \textbf{mAP} & \textbf{NDS} & \textbf{mAP} & \textbf{NDS} & \textbf{mAP} & \textbf{NDS}
        \\\midrule\midrule
        \multicolumn{8}{c}{\textbf{Backbone:} {VoxelNet \cite{zhou2018voxelnet} + CenterPoint \cite{yin2021centerpoint}}}
        \\\midrule
        \textcolor{gray}{Random} & \textcolor{gray}{-} & \textcolor{gray}{${38.0}$} & \textcolor{gray}{${44.3}$} & \textcolor{gray}{${46.9}$} & \textcolor{gray}{${55.5}$} & \textcolor{gray}{${50.2}$} & \textcolor{gray}{${59.7}$}
        \\
        PointContrast \cite{xie2020pointcontrast} & ECCV'20 & $39.8$ & $45.1$ & $47.7$ & $56.0$ & - & -
        \\
        GCC-3D \cite{liang2021gcc-3d} & ICCV'21 & $41.1$ & $46.8$ & $48.4$ & $56.7$ & - & -
        \\
        SLidR \cite{sautier2022slidr} & CVPR'22 & $43.3$ & $52.4$ & $47.5$ & $56.8$ & $50.4$ & $59.9$
        \\
        TriCC \cite{pang2023tricc} & CVPR'23 & $44.6$ & \hlred{$\mathbf{54.4}$} & $48.9$ & $58.1$ & $50.9$ & $60.3$
        \\
        CSC \cite{chen2024csc} & CVPR'24 & \hlred{$\mathbf{45.3}$} & $54.2$ & \hlred{$\mathbf{49.3}$} & \hlred{$\mathbf{58.3}$} & \hlred{$\mathbf{51.9}$} & \hlred{$\mathbf{61.3}$}
        \\
        ScaLR \cite{puy2024scalr} & CVPR'24 & $44.3$ & $53.3$ & $48.2$ & $57.1$ & $50.7$ & $60.8$
        \\
        \textbf{LiMA} & \textbf{Ours} & \hlgreen{$\mathbf{46.5}$} & \hlgreen{$\mathbf{56.4}$} & \hlgreen{$\mathbf{50.1}$} & \hlgreen{$\mathbf{59.6}$} & \hlgreen{$\mathbf{52.3}$} & \hlgreen{$\mathbf{62.3}$}
        \\\midrule
        \multicolumn{8}{c}{\textbf{Backbone:} {VoxelNet \cite{zhou2018voxelnet} + SECOND \cite{yan2018second}}}
        \\\midrule
        \textcolor{gray}{Random} & \textcolor{gray}{-} & \textcolor{gray}{${35.8}$} & \textcolor{gray}{${45.9}$} & \textcolor{gray}{${39.0}$} & \textcolor{gray}{${51.2}$} & \textcolor{gray}{${43.1}$} & \textcolor{gray}{${55.7}$}
        \\
        SLidR \cite{sautier2022slidr} & CVPR'22 & $36.6$ & $48.1$ & $39.8$ & $52.1$ & $44.2$ & $56.3$
        \\
        TriCC \cite{pang2023tricc} & CVPR'23 & $37.8$ & \hlred{$\mathbf{50.0}$} & $41.4$ & $53.5$ & $45.5$ & $57.7$
        \\
        CSC \cite{chen2024csc} & CVPR'24 & \hlred{$\mathbf{38.2}$} & $49.4$ & \hlred{$\mathbf{42.5}$} & \hlred{$\mathbf{54.8}$} & \hlred{$\mathbf{45.6}$} & $58.1$
        \\
        ScaLR \cite{puy2024scalr} & CVPR'24 & $37.3$ & $48.7$ & $41.4$ & $53.5$ & $45.5$ & \hlred{$\mathbf{58.6}$}
        \\
        \textbf{LiMA} & \textbf{Ours} & \hlgreen{$\mathbf{39.4}$} & \hlgreen{$\mathbf{50.1}$} & \hlgreen{$\mathbf{43.2}$} & \hlgreen{$\mathbf{55.3}$} & \hlgreen{$\mathbf{46.0}$} & \hlgreen{$\mathbf{59.5}$}
        \\\bottomrule
    \end{tabular}}
    \vspace{-0.15cm}
\end{table}

\section{Experiments}
\label{sec:experiments}

\subsection{Experimental Settings}

\noindent\textbf{Datasets.}
Following standard practice \cite{sautier2022slidr,liu2023seal,chen2024csc}, we use \textit{nuScenes} \cite{caesar2020nuscenes} for model pretraining. For downstream evaluation, we evaluate our approach on a diverse set of datasets, including \textit{nuScenes} \cite{caesar2020nuscenes, fong2022panoptic-nuscenes}, \textit{SemanticKITTI} \cite{behley2019semantickitti}, \textit{ScribbleKITTI} \cite{unal2022scribblekitti}, \textit{Waymo Open} \cite{sun2020waymo}, \textit{RELLIS-3D} \cite{jiang2021rellis-3d}, \textit{SemanticPOSS} \cite{pan2020semanticposs}, \textit{SemanticSTF} \cite{xiao2023semanticstf}, \textit{SynLiDAR} \cite{xiao2022synlidar}, \textit{DAPS-3D} \cite{klokov2023daps3d}, \textit{Synth4D} \cite{saltori2022synth4d}, and \textit{nuScenes-C} \cite{kong2023robo3d}.

\noindent\textbf{Implementation Details.} 
Our experiments are conducted using \textit{MMDetection3D}~\cite{mmdet3d}. We adopt \textit{MinkUNet} \cite{choy2019minkunet} for LiDAR semantic segmentation and \textit{VoxelNet} \cite{zhou2018voxelnet} for 3D object detection. The 2D backbone is \textit{ViT} \cite{dosovitskiy2021vit} (small, base, large), pretrained with \textit{DINOv2} \cite{oquab2023dinov2}. Pretraining is performed for $50$ epochs on $8$ GPUs with a batch size of $2$ per GPU, using the AdamW optimizer \cite{loshchilov2017adamw} (initial learning rate $0.01$) and the OneCycle scheduler \cite{smith2019onecycle}. To capture long-term temporal dependencies, we use $6$ consecutive frames. During downstream fine-tuning, task-specific heads are trained with a learning rate $10\times$ higher than the backbone. All evaluations are performed without test-time augmentations for fair comparison. We report mean Intersection-over-Union (mIoU) for segmentation, mean Corruption Error (mCE) and mean Resilience Rate (mRR) for robustness, and mean Average Precision (mAP) and nuScenes Detection Score (NDS) for detection.

\subsection{Comparative Study}

\noindent\textbf{In-Domain Fine-Tuning.}  We evaluate LiMA against state-of-the-art pretraining methods using both Linear Probing (LP) and few-shot fine-tuning on the \textit{nuScenes} dataset \cite{fong2022panoptic-nuscenes}. As shown in \cref{tab:benchmark}, LiMA demonstrates significant improvements under limited annotation settings (\eg, $1\%$, $5\%$, and $10\%$). Specifically, when distilled from ViT-S to ViT-L, LiMA achieves mIoU scores of $54.76\%$, $56.65\%$, and $56.65\%$ in the LP, yielding gains of $5.08\%$, $4.75\%$, and $4.90\%$ over the baseline \cite{puy2024scalr}. Moreover, LiMA consistently outperforms prior methods across all fine-tuning settings, achieving a $2\%$ to $3\%$ mIoU improvement. These results underscore the effectiveness of long-term distillation in capturing temporal dynamics and maintaining view consistency, leading to more robust feature representations.

\noindent\textbf{Cross-Domain Generalization.} To assess the scalability of LiMA, we conduct a comprehensive evaluation across nine diverse 3D semantic segmentation datasets, each representing distinct driving scenarios. As shown in \cref{tab:benchmark} and \cref{tab:multiple_dataset}, LiMA consistently outperforms baseline methods across all settings. These results highlight our strong feature representation capabilities, demonstrating its adaptability to varying data distributions and its ability to maintain high performance across heterogeneous environments.

\noindent\textbf{Out-of-Distribution Robustness.} Evaluating a model's resilience to out-of-distribution scenarios is critical for assessing its robustness. As shown in \cref{tab:robo3d}, we conduct a comprehensive robustness evaluation on the \textit{nuScenes-C} dataset from Robo3D \cite{kong2023robo3d}. LiMA consistently outperforms recent pretraining methods across most corruption types, highlighting its enhanced ability to preserve performance under adverse conditions. This robustness is particularly important for real-world applications, where models must adapt to diverse and unpredictable environments.

\noindent\textbf{Fine-Tuning for 3D Object Detection.} To evaluate the generalization capability of LiMA, we fine-tune our pretrained model within the SECOND \cite{yan2018second} and CenterPoint \cite{yin2021centerpoint} detection frameworks. As shown in \cref{tab:detection}, LiMA yields significant improvements in detection accuracy over state-of-the-art methods. These results demonstrate the effectiveness of LiMA in transferring learned features across different downstream tasks, further validating its robustness and versatility in diverse and challenging scenarios.

\begin{figure}[t]
    \centering
    \includegraphics[width=\linewidth]{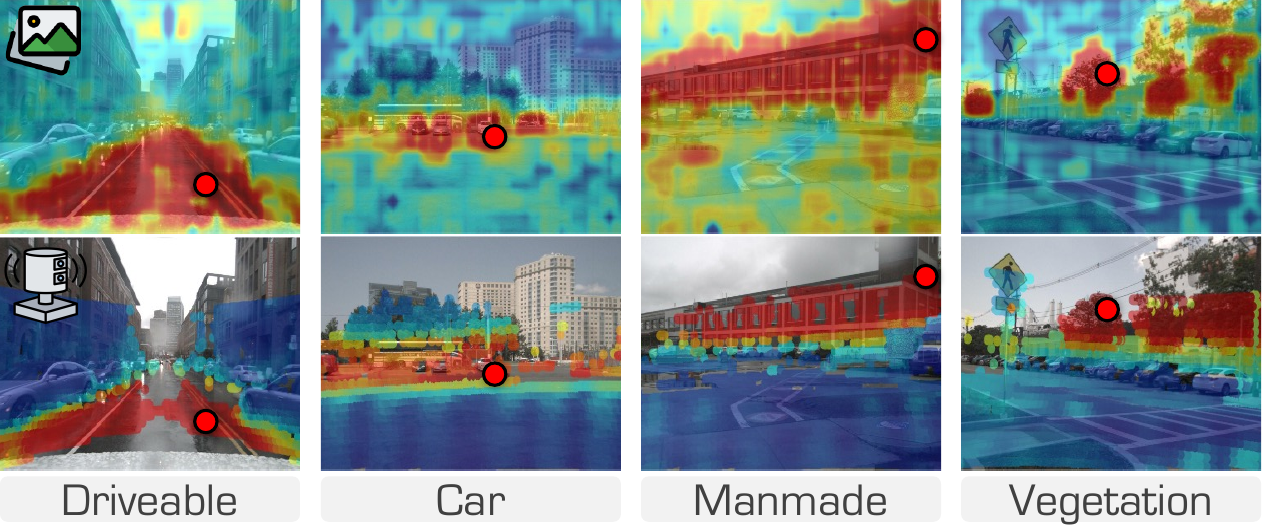}
    \vspace{-0.7cm}
    \caption{\textbf{Cosine similarity} between a query point (marked as \textcolor{term_red}{\textbf{red}} dot) and: (1) image features, and (2) LiDAR features projected onto the image. Colors range from \textcolor{term_red}{\textbf{red}} (indicating \textcolor{term_red}{\textbf{high}} similarity) to \textcolor{term_blue}{\textbf{blue}} (indicating \textcolor{term_blue}{\textbf{low}} similarity). Best viewed in colors.}
    \label{fig:similarity}
    \vspace{-0.2cm}
\end{figure}

\begin{figure}[t]
    \centering
    \includegraphics[width=\linewidth]{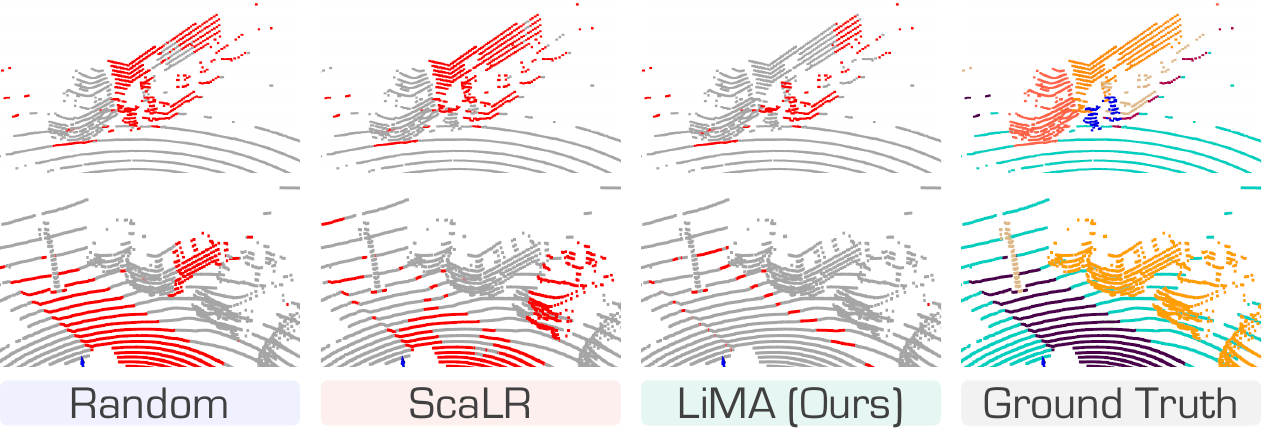}
    \vspace{-0.7cm}
    \caption{\textbf{Qualitative assessments} of state-of-the-art methods, pretrained on \textit{nuScenes} \cite{caesar2020nuscenes} and fine-tuned on \textit{nuScenes} \cite{fong2022panoptic-nuscenes} with $1\%$ annotations. The error maps depict \textcolor{gray}{\textbf{correct}} and \textcolor{term_red}{\textbf{incorrect}} predictions in \textcolor{gray}{\textbf{gray}} and \textcolor{term_red}{\textbf{red}}, respectively. Best viewed in colors.}
    \label{fig:seg}
    \vspace{-0.2cm}
\end{figure}

\begin{figure}[t]
    \centering
    \includegraphics[width=\linewidth]{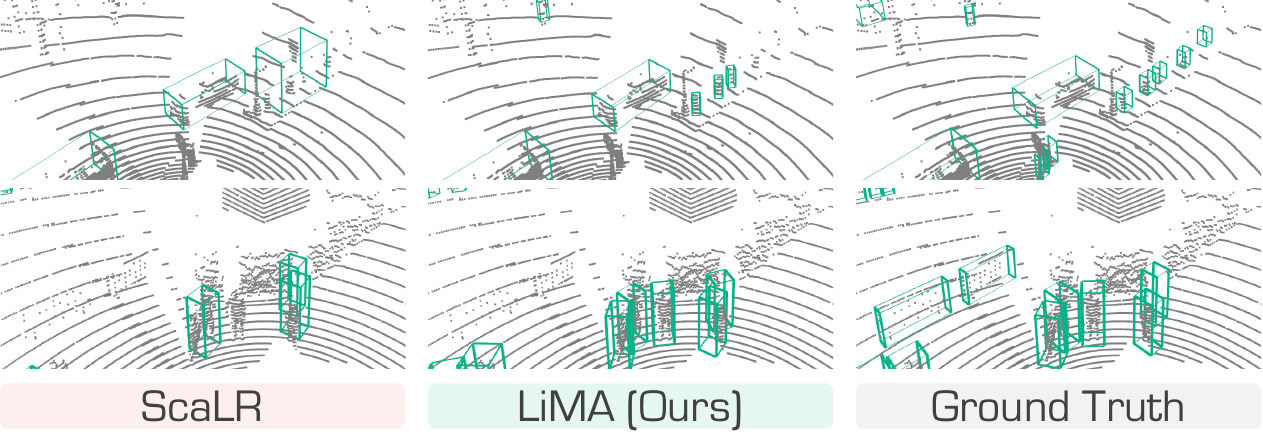}
    \vspace{-0.7cm}
    \caption{\textbf{Visual comparisons} from the 3D object detection task, where methods are pretrained on \textit{nuScenes} \cite{caesar2020nuscenes} and fine-tuned on \textit{nuScenes} \cite{caesar2020nuscenes} with $5\%$ annotations. Best viewed in colors.}
    \label{fig:det}
    \vspace{-0.2cm}
\end{figure}

\noindent\textbf{Qualitative Results.} \cref{fig:similarity} visualizes the similarity between a query point and the pretrained 2D image backbone, as well as other LiDAR points. The results demonstrate that long-term information ensures semantic coherence during pretraining, leading to more effective feature learning. Additionally, qualitative segmentation and detection results, presented in \cref{fig:seg} and \cref{fig:det}, respectively, demonstrate that LiMA produces more precise predictions, particularly for challenging object classes and dynamic scenes. By effectively integrating spatial and temporal cues, LiMA mitigates misclassifications and enhances object localization, further validating its robustness in real-world scenarios.

\begin{table}[t]
    \centering
    \caption{Ablation study of \textbf{each component} in LiMA. All variants use ViT-B \cite{dosovitskiy2021vit} for distillation. \textbf{CVA}: Cross-view aggregation. \textbf{LTFP}: Long-term feature propagation. \textbf{CSMA}: Cross-sequence memory alignment. All scores are given in percentage (\%).}
    \vspace{-0.3cm}
    \label{tab:architecture}
    \resizebox{\linewidth}{!}{
    \begin{tabular}{c|ccc|ccc|c}
        \toprule
        \multirow{2}{*}{\textbf{\#}} & \textbf{CVA} & \textbf{LTFP} & \textbf{CSMA} & \multicolumn{3}{c|}{\textbf{nuScenes}} & \textbf{KITTI}
        \\
        & {\small(\cref{sec:cva})} & {\small(\cref{sec:term})} & {\small(\cref{sec:cross-sequence})} & \textbf{LP} & \textbf{1\%} & \textbf{5\%} & \textbf{1\%}
        \\\midrule\midrule
        (a) & \textcolor{term_red}{\xmark} & \textcolor{term_red}{\xmark} & \textcolor{term_red}{\xmark} & \textcolor{gray}{$51.90$} & \textcolor{gray}{$48.90$} & \textcolor{gray}{$57.69$} & \textcolor{gray}{$47.77$}
        \\\midrule
        (b) & \textcolor{term_green}{\cmark} & \textcolor{term_red}{\xmark} & \textcolor{term_red}{\xmark} & $53.72$ & $49.52$ & $58.34$ & $48.75$
        \\
        (c) & \textcolor{term_green}{\cmark} & \textcolor{term_green}{\cmark} & \textcolor{term_red}{\xmark} & $55.26$ & $50.43$ & $60.03$ & $49.31$
        \\
        (d) & \textcolor{term_green}{\cmark} & \textcolor{term_green}{\cmark} & \textcolor{term_green}{\cmark} & \hlgreen{$\mathbf{56.65}$} & \hlgreen{$\mathbf{51.29}$} & \hlgreen{$\mathbf{61.11}$} & \hlgreen{$\mathbf{50.44}$}
        \\\bottomrule
    \end{tabular}}
    \vspace{-0.35cm}
\end{table}

\subsection{Ablation Study}

\noindent\textbf{Component Analysis.} \cref{tab:architecture} presents an ablation study evaluating the three key components of LiMA, each contributing to enhanced LiDAR representation learning. First, cross-view aggregation improves fine-tuning mIoU by $0.8\%$ and LP by $1.82\%$ by unifying overlapping regions across viewpoints. This ensures consistent point-wise representations and mitigates ``optimization conflicts'' arising from inconsistent feature learning across views. Second, the long-term feature propagation module facilitates motion pattern learning by leveraging historical image features, leading to an improvement of over $1\%$ in fine-tuning performance and a $1.54\%$ gain in LP. This highlights the importance of incorporating past observations to refine feature representations and enhance temporal consistency. Finally, to improve robustness and adaptability in complex, diverse environments, we introduce the cross-sequence memory alignment. This strategy achieves a $0.9\%$ mIoU gain in fine-tuning and a $1.39\%$ improvement in LP, demonstrating its effectiveness in enabling better generalization across driving scenarios.

\begin{table}[t]
    \centering
    \caption{Ablation study on \textbf{aggregation methods} in the cross-view aggregation module. All scores are given in percentage (\%).}
    \vspace{-0.3cm}
    \label{tab:aggregation}
    \resizebox{\linewidth}{!}{
    \begin{tabular}{c|c|ccc|c|c}
        \toprule
        \multirow{2}{*}{\textbf{\#}} & \multirow{2}{*}{\textbf{Aggregation Strategy}} & \multicolumn{3}{c|}{\textbf{nuScenes}} & \textbf{KITTI} & \textbf{Waymo}
        \\
        & & \textbf{LP} & \textbf{1\%} & \textbf{5\%} & \textbf{1\%} & \textbf{1\%}
        \\\midrule\midrule
        (a) & \textcolor{gray}{None} & \textcolor{gray}{$54.64$} & \textcolor{gray}{$49.34$} & \textcolor{gray}{$59.46$} & \textcolor{gray}{$48.02$} & \textcolor{gray}{$50.76$}
        \\\midrule
        (b) & Maximum & $55.93$ & $50.02$ & $60.23$ & $49.43$ & $50.91$
        \\
        (c) & Average & \hlgreen{$\mathbf{56.65}$} & \hlgreen{$\mathbf{51.29}$} & \hlgreen{$\mathbf{61.11}$} & \hlgreen{$\mathbf{50.44}$} & \hlgreen{$\mathbf{51.35}$}
        \\
        (d) & Attention Module & $55.02$ & $49.87$ & $59.92$ & $49.32$ & $50.23$
        \\\bottomrule
    \end{tabular}}
    \vspace{-0.3cm}
\end{table}

\begin{table}[t]
    \centering
    \caption{Ablation study on the \textbf{effect of varying frames} for long-term information aggregation. Training time and memory usage are measured with a batch size of $2$ on eight A800 GPUs.}
    \vspace{-0.3cm}
    \label{tab:time_step}
    \resizebox{\linewidth}{!}{
    \begin{tabular}{r|c|c|c|cc|c}
        \toprule
        \multirow{2}{*}{\textbf{Method}} & \multirow{2}{*}{\textbf{Frames}} & \textbf{Training Time} &
        \textbf{Memory} & \multicolumn{2}{c|}{\textbf{nuScenes}} & \textbf{KITTI}
        \\
        & & \textbf{(Hours)} & \textbf{(GB)} & \textbf{LP} & \textbf{1\%} & \textbf{1\%}
        \\\midrule\midrule
        ScaLR~\cite{puy2024scalr} & $1$ & $\sim 10.1$ & $12.43$ & $51.90$ & $48.90$ & $47.77$
        \\\midrule
        \multirow{7}{*}{\textbf{LiMA (Ours)}} & $2$ & $\sim 14.7$ & $18.23$ & $53.34$ & $49.14$ & $48.27$
        \\
        & $3$ & $\sim 15.3$ & $20.67$ & $54.52$ & $49.75$ & $48.92$
        \\
        & $4$ & $\sim 16.1$ & $23.19$ & $55.65$ & $50.29$ & $49.44$
        \\
        & $5$ & $\sim 17.0$ & $26.59$ & $56.03$ & $50.95$ & $50.32$
        \\
        & $6$ & $\sim 17.9$ & $29.07$ & \hlgreen{$\mathbf{56.65}$} & \hlgreen{$\mathbf{51.29}$} & $50.44$
        \\
        & $7$ & $\sim 18.7$ & $33.55$ & $55.37$ & $50.91$ & $51.00$
        \\
        & $8$ & $\sim 19.5$ & $36.27$ & $54.97$ & $49.36$ & \hlgreen{$\mathbf{51.78}$}
        \\\midrule
        Seal~\cite{liu2023seal} & $2$ & $\sim 27.3$ & $20.92$ & $46.59$ & $45.98$ & $47.24$
        \\
        SuperFlow~\cite{xu2024superflow} & $3$ & $\sim 30.7$ & $23.65$ & $47.66$ & $48.09$ & $48.40$
        \\\bottomrule
    \end{tabular}}
    \vspace{-0.2cm}
\end{table}

\noindent\textbf{Cross-View Feature Aggregation Strategy.} Cross-view feature aggregation is a crucial module for unifying multi-view features. In this ablation study, we evaluate different aggregation methods, including max-pooling, average-pooling, and attention-based approaches as shown in \cref{tab:aggregation}. The results indicate that average-pooling achieves the best performance, as it effectively balances complementary visual features across different views. In contrast, max-pooling captures prominent features but suppresses finer details, leading to suboptimal results. Meanwhile, the attention-based method introduces additional learnable parameters, increasing optimization complexity.

\noindent\textbf{Effects of Memory Bank Frames.} We conduct an ablation study to evaluate the impact of memory bank size on long-term information aggregation, as shown in \cref{tab:time_step}. Increasing the number of stored frames generally improves performance by providing a richer temporal context compared to the spatial alignment baseline \cite{puy2024scalr}. However, beyond an optimal threshold, excessive historical frames introduce feature misalignment due to calibration errors and temporal drift, leading to inconsistencies in the aggregated representations. This underscores a trade-off between leveraging long-term dependencies and maintaining feature coherence, as accumulating too many past frames may introduce noise from moving scenes rather than beneficial information.

\noindent\textbf{Efficacy Analysis.} As shown in \cref{tab:time_step}, we observe the following: \textbf{(1)} Compared to prior temporal contrastive methods \cite{liu2023seal,xu2024superflow}, LiMA achieves significantly higher pretraining efficiency, requiring no more than $20$ hours even when propagating across $8$ frames, whereas prior contrastive methods exceed $24$ hours of computation with limited frames. \textbf{(2)} Despite extended temporal modeling, long-term propagation does not substantially increase pretraining time or memory consumption. This efficiency gain stems from the memory bank mechanism, which efficiently stores and retrieves informative features, effectively eliminating redundant computations and optimizing resource utilization.

\begin{figure}[t]
    \centering
    \begin{subfigure}[h]{0.48\linewidth}
        \centering
        \includegraphics[width=\linewidth]{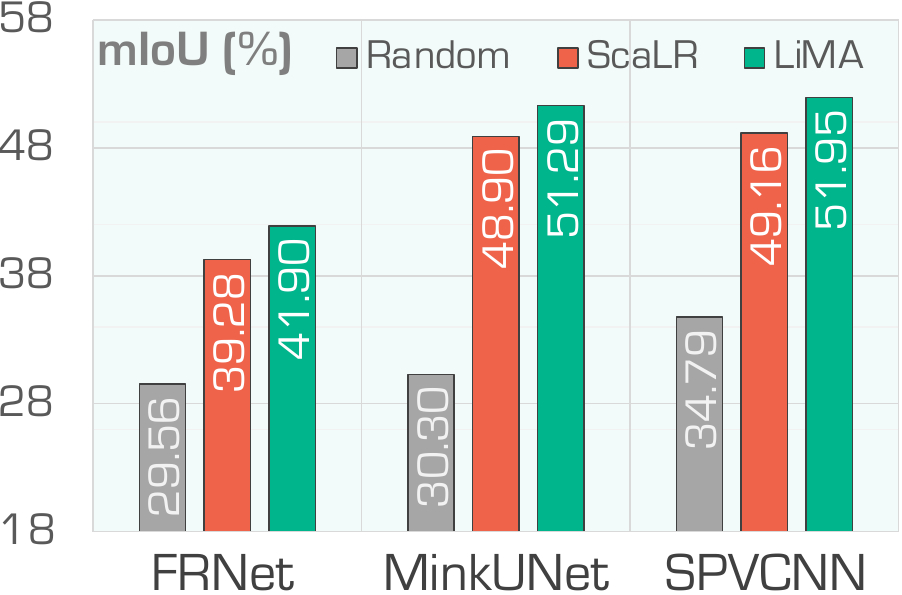}
        \caption{nuScenes ($1$\%)}
    \end{subfigure}
    ~
    \begin{subfigure}[h]{0.48\linewidth}
        \centering
        \includegraphics[width=\linewidth]{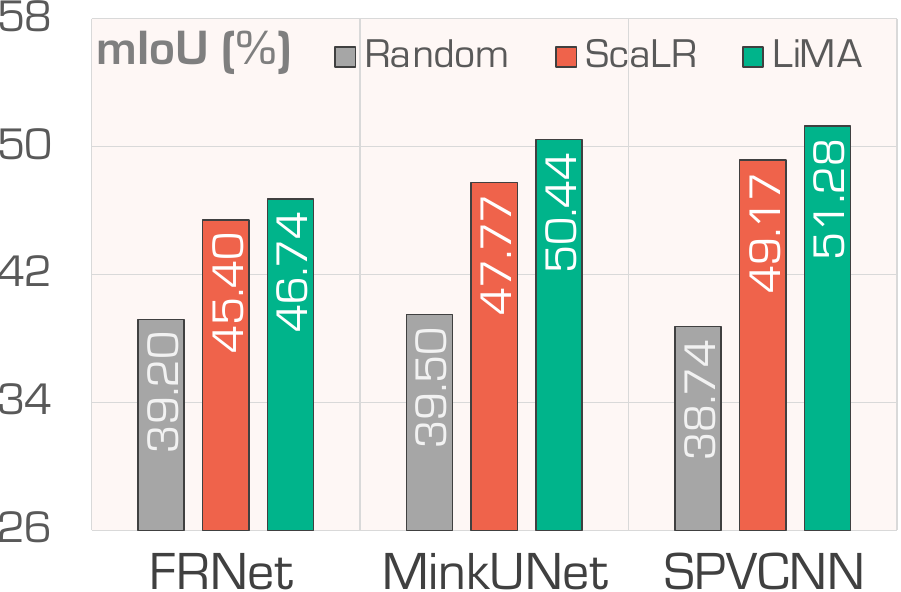}
        \caption{SemanticKITTI ($1$\%)}
    \end{subfigure}
    \vspace{-0.3cm}
    \caption{Ablation study on \textbf{different backbones} for downstream tasks. The backbones are initialized with random weights, ScaLR \cite{puy2024scalr}, and LiMA, respectively, and fine-tuned on the nuScenes \cite{fong2022panoptic-nuscenes} and SemanticKITTI \cite{behley2019semantickitti} datasets using $1\%$ annotations.}
    \label{fig:backbone}
    \vspace{-0.3cm}
\end{figure}

\noindent\textbf{Representations from 3D Backbones.} To evaluate the adaptability of LiMA across different 3D architectures, we replace the 3D backbone with SPVCNN \cite{tang2020spvcnn} and FRNet \cite{xu2025frnet}. The results in \cref{fig:backbone} demonstrate that LiMA consistently outperforms the baseline across diverse representations, underscoring its robustness and flexibility. This highlights LiMA's ability to effectively integrate with various 3D backbones while maintaining superior performance.

\section{Conclusion}
\label{sec:conclusion}

This work proposes image-to-\textbf{Li}DAR \textbf{M}emory \textbf{A}ggregation (\textbf{LiMA}), a new data pretraining framework that enhances the robustness and adaptability of LiDAR-based perception systems in dynamic scenarios. LiMA effectively captures temporal dynamics and diversifies training data through three key designs: cross-view aggregation, long-term feature propagation, and cross-sequence memory alignment. These modules enable scalable and effective pretraining. Extensive experiments across various tasks and datasets demonstrate the superiority of LiMA, while evaluations on different 3D backbones highlight its strong generalization capabilities. Our findings position LiMA as a promising approach for developing powerful 3D foundation models.


\section*{Acknowledgments}

This work was supported in part by the Natural Science Foundation of China under Grants U24B20155 and U21B2044, and the Key Research and Development Program of Jiangsu Province under Grant BE2023016-3. This work was also supported in part by the National Natural Science Foundation of China under Grant 62402245, the Natural Science Foundation of Jiangsu Province under Grant BK20240644, and the Natural Science Foundation of the Jiangsu Higher Education Institutions of China under Grant 24KJB520023.

\section*{Appendix}
\startcontents[appendices]
\printcontents[appendices]{l}{1}{\setcounter{tocdepth}{3}}

\section{Additional Implementation Details}
\label{sec:supp_implement}

In this section, we provide a comprehensive overview of the datasets, hyperparameter settings, and the training and evaluation protocols employed in our experiments.

\subsection{Datasets}

In this work, we conduct extensive experiments across varying driving datasets, covering both urban and campus driving environments, various LiDAR sensor configurations, and a wide range of static and dynamic object distributions. Our datasets span real-world and synthetic scenarios, enabling a comprehensive evaluation of model performance across different environmental conditions. Below, we provide a detailed overview of each dataset used in this work:

\begin{itemize}
    \item \textbf{nuScenes \cite{fong2022panoptic-nuscenes}:} A large-scale, multi-modality dataset designed for autonomous driving, featuring a 32-beam LiDAR and six cameras capturing diverse urban driving scenes. The dataset consists of $1,000$ sequences collected in Boston and Singapore, with a standard split of $700$ training, $150$ validation, and $150$ testing scenes. For pretraining, we adopt the SLidR \cite{sautier2022slidr} protocol, partitioning the training set into a mini-train/val split ($600$ training, $100$ validation). To systematically analyze model performance under different data availability levels, we construct sub-training sets by uniformly sampling $1\%$, $5\%$, $10\%$, $25\%$, and $100\%$ of the training data for fine-tuning.

    \item \textbf{SemanticKITTI \cite{behley2019semantickitti}:} A large-scale dataset tailored for semantic scene understanding in autonomous driving, featuring high-resolution 3D point clouds captured by a 64-beam LiDAR sensor. It provides dense point-wise annotations for all $22$ sequences from the KITTI Odometry Benchmark \cite{geiger2012kitti}, covering a diverse range of driving environments, including urban streets, highways, and residential areas. This dataset serves as a benchmark for evaluating semantic segmentation models in real-world autonomous driving scenarios. To systematically assess model performance in data-efficient settings, we construct a $1\%$ sub-training set by uniformly sampling training scans, enabling experiments in low-data regimes.

    \item \textbf{Waymo Open \cite{sun2020waymo}:} A large-scale dataset collected from real-world autonomous driving scenarios, comprising $1,150$ driving sequences -- $798$ for training, $202$ for validation, and $150$ for testing. The dataset is captured using a multi-LiDAR setup, consisting of one mid-range and four short-range LiDAR sensors, enabling the collection of dense and diverse point clouds across different traffic conditions. To assess model performance in data-efficient settings, we construct a $1\%$ sub-training set through uniform sampling of the training scans.

    \item \textbf{ScribbleKITTI \cite{unal2022scribblekitti}:} A weakly labeled variant of the SemanticKITTI dataset \cite{behley2019semantickitti}, where annotations are provided as sparse line scribbles instead of dense point-wise labels. This dataset is specifically designed to advance research in weakly supervised learning for semantic segmentation in autonomous driving, reducing the reliance on expensive manual annotations. By leveraging its sparse annotations, models can be trained to learn robust semantic representations with limited supervision. In this work, we construct $1\%$ and $10\%$ sub-training sets through uniform sampling to systematically investigate the effectiveness of weak supervision in data-efficient learning scenarios.

    \item \textbf{RELLIS-3D \cite{jiang2021rellis-3d}:} A multimodal dataset collected in off-road environments at the Rellis Campus of Texas A\&M University, specifically designed to support research in unstructured terrain scenarios. This dataset contains diverse and complex scenes with vegetation, rough surfaces, and varying elevations, posing challenges distinct from structured urban driving. In this work, we construct $1\%$ and $10\%$ sub-training sets to systematically evaluate model generalization in non-urban environments.

    \item \textbf{SemanticPOSS \cite{pan2020semanticposs}:} A small-scale dataset collected in off-road environments at Peking University, designed to evaluate the performance of autonomous perception systems in complex, unstructured scenes. It captures diverse scenarios involving both static and dynamic objects, featuring dense point clouds that pose challenges distinct from structured urban settings. The dataset comprises $6$ sequences, providing a compact yet challenging benchmark for semantic segmentation. In this work, we utilize sequences $00$ and $01$ to form a half sub-training set, while sequences $00$ to $05$ (excluding $02$ for validation) constitute the full training set, ensuring a systematic evaluation of model performance in off-road conditions.

    \item \textbf{SemanticSTF \cite{xiao2023semanticstf}:} A dataset specifically designed to evaluate the robustness of LiDAR-based perception models under adverse weather conditions. It includes four challenging scenarios: \textit{``snowy''}, \textit{``dense fog''}, \textit{``light fog''}, and \textit{``rain''}, simulating real-world environmental variations that can significantly impact LiDAR sensor performance. Captured across diverse outdoor settings, this dataset serves as a critical benchmark for assessing the degradation of semantic segmentation models when exposed to extreme weather conditions. To systematically analyze model performance in data-efficient learning settings, we construct both half and full sub-training sets by uniformly sampling training scans.

    \item \textbf{SynLiDAR \cite{xiao2022synlidar}:} A large-scale synthetic dataset generated using Unreal Engine 4, designed to simulate diverse LiDAR perception scenarios in unstructured virtual environments. It consists of $13$ sequences, totaling $198,396$ scans, providing high-quality point clouds that closely mimic real-world sensor data while eliminating the need for costly manual annotations. The dataset covers various terrains and object distributions, making it a valuable resource for studying domain adaptation and generalization in LiDAR-based perception models. In this work, we construct $1\%$ and $10\%$ sub-training sets through uniform sampling to systematically evaluate model performance.

    \item \textbf{DAPS-3D \cite{klokov2023daps3d}:} A dataset comprising two subsets: DAPS-1, a semi-synthetic collection featuring large-scale 3D scenes, and DAPS-2, which contains real-world LiDAR scans recorded by a cleaning robot operating in VDNH Park, Moscow. This dataset is specifically designed to facilitate research on domain adaptation and semi-synthetic training paradigms by bridging the gap between synthetic and real-world data. In this work, we extract training scans from the sequence ``38-18\_7\_72\_90'' within the DAPS-1 subset and construct both half and full sub-training sets to evaluate model performance.

    \item \textbf{Synth4D \cite{saltori2022synth4d}:} A synthetic dataset generated using the CARLA simulator, designed to replicate real-world driving conditions with Velodyne LiDAR sensors. It comprises two subsets: Synth4D-KITTI and Synth4D-nuScenes, each crafted to closely resemble their respective real-world counterparts. In this work, we utilize the Synth4D-nuScenes subset and construct $1\%$ and $10\%$ sub-training sets by uniformly sampling training scans to assess model performance in low-data regimes.

    \item \textbf{nuScenes-C \cite{kong2023robo3d}:} An extension of the nuScenes dataset \cite{caesar2020nuscenes}, designed to evaluate the robustness of LiDAR-based perception models under challenging environmental conditions. The dataset introduces eight types of synthetic corruptions -- \textit{``fog''}, \textit{``wet ground''}, \textit{``snow''}, \textit{``motion blur''}, \textit{``beam missing''}, \textit{``crosstalk''}, \textit{``implement echo''}, and \textit{``cross-sensor''} -- each with three severity levels: easy, moderate, and hard. These corruptions simulate the real-world effects of sensor degradation and environmental disturbances, such as bad weather or sensor malfunctions, that can degrade perception performance. This dataset serves as a benchmark to assess model resilience against these distortions, providing a comprehensive tool to evaluate the reliability and robustness of perception systems, particularly in autonomous driving applications.
\end{itemize}

\subsection{Training Configurations}

\noindent\textbf{Image Preprocessing.} For image-based inputs, we apply standard data augmentation techniques. Specifically, we apply horizontal flipping with a $50\%$ probability and resize all input images to a fixed resolution of $448 \times 224$ pixels to ensure consistency across different datasets. Following ScaLR \cite{puy2024scalr}, in this work, we do not utilize any Vision Foundation Models (VFMs) to generate superpixels in this work.

\noindent\textbf{Point Cloud Preprocessing.} For LiDAR point clouds, we employ a set of geometric transformations to enhance data diversity during training. We apply random rotations around the $z$-axis within the range of $[-180^{\circ}, 180^{\circ}]$, which accounts for minor variations in sensor orientation. Additionally, we randomly flip the point cloud along the $x$-axis and $y$-axis with a probability of $50\%$ each. To introduce scale invariance, we apply a random scaling factor sampled uniformly from the range $[0.95, 1.05]$. Finally, we voxelize the point cloud using a cylindrical partitioning strategy with a voxel resolution of $0.1$ meters.

\noindent\textbf{Optimization.} We employ the AdamW optimizer \cite{loshchilov2017adamw} for training, with an initial learning rate of $0.01$. To dynamically adjust the learning rate, we adopt the OneCycle learning rate scheduler \cite{smith2019onecycle}. The per-GPU batch size is set to $2$, resulting in a total batch size of $16$ across $8$ GPUs. For downstream fine-tuning, we apply a differentiated learning rate strategy. The backbone network is trained with a lower learning rate, while the task-specific head is updated with a learning rate that is $10\times$ higher to facilitate rapid adaptation to new tasks. Fine-tuning experiments are conducted across multiple data regimes, systematically evaluating model performance under varying levels of data availability.

\subsection{Evaluation Configurations}

This section outlines the evaluation metrics used to assess model performance in 3D semantic segmentation, robustness evaluation, and 3D object detection. Each metric is designed to quantify different aspects of model effectiveness.

\noindent\textbf{3D Semantic Segmentation.} We follow standard evaluation practices by reporting the Intersection-over-Union (IoU) for each category and the mean IoU (mIoU) across all categories. The IoU for a given class $i$ is defined as:
\begin{equation}
    \label{equ:iou}
    \text{IoU}_{i} = \frac{\text{TP}_{i}}{\text{TP}_{i} + \text{FP}_{i} + \text{FN}_{i}}~,
\end{equation}
where $\text{TP}_{i}$ (True Positives) denotes correctly classified points, $\text{FP}_{i}$ (False Positives) represents incorrectly predicted points, and $\text{FN}_{i}$ (False Negatives) accounts for misclassified ground-truth points. Higher IoU values indicate better segmentation accuracy. The mean IoU (mIoU) is computed as the average IoU across all classes:
\begin{equation}
    \label{equ:miou}
    \text{mIoU} = \frac{1}{|\mathbb{C}|} \sum_{i \in \mathbb{C}} \text{IoU}_{i}~,
\end{equation}
where $\mathbb{C}$ represents the set of all semantic categories. mIoU provides a holistic measure of overall segmentation quality.

\noindent\textbf{Robustness Evaluation.} To evaluate model robustness against real-world perturbations, we adopt the Corruption Error (CE) and Resilience Rate (RR) metrics, following the Robo3D \cite{kong2023robo3d} evaluation protocol. These metrics assess performance under various corruption types, including sensor noise, adverse weather conditions, and motion blur.

\begin{itemize}
    \item \textbf{Corruption Error (CE).} CE quantifies the model's performance degradation under corruption type $i$, defined as:
    \begin{equation}
        \label{equ:ce}
        \text{CE}_{i}=\frac{\sum_{j=1}^{3}(1-\text{IoU}_{i,j})}{\sum_{j=1}^{3}(1-\text{IoU}_{i,j}^{\text{base}})}~,
    \end{equation}
    where $\text{IoU}_{i,j}$ represents the mIoU under corruption type $i$ at severity level $j$, and $\text{IoU}_{i,j}^{\text{base}}$ is the mIoU of a baseline model under the same conditions. Lower CE values indicate greater robustness, meaning the model maintains performance even under significant corruption.

    \item \textbf{Resilience Rate (RR).} RR measures the model's ability to recover from corruptions, computed as:
    \begin{equation}
        \label{equ:rr}
        \text{RR}_{i}=\frac{\sum_{j=1}^{3}
        \text{IoU}_{i,j}}{3 \times \text{IoU}^{\text{clean}}}~,
    \end{equation}
    where $\text{IoU}^{\text{clean}}$ represents the mIoU on the ``clean'' validation set, which serves as a reference for the model's performance under ideal conditions. A higher RR signifies greater resilience to external disturbances.
\end{itemize}
Additionally, we report the mean Corruption Error (mCE) and mean Resilience Rate (mRR) across all corruption types to summarize overall robustness.

\noindent\textbf{3D Object Detection.} We evaluate 3D object detection performance using the mean Average Precision (mAP) and the nuScenes Detection Score (NDS), following nuScenes \cite{caesar2020nuscenes}.

\begin{itemize}
    \item \textbf{Mean Average Precision (mAP).} mAP is computed by averaging the Average Precision (AP) across object classes and distance-based matching thresholds:
    \begin{equation}
        \label{equ:mAP}
        \text{mAP} = \frac{1}{|\mathbb{C}| |\mathbb{D}|} \sum_{c\in\mathbb{C}}\sum_{d\in\mathbb{D}} \text{AP}_{c,d}~,
    \end{equation}
    where $\mathbb{C}$ denotes the set of object categories, and $\mathbb{D} = \{0.5, 1, 2, 4\}$ represents different distance thresholds used for AP computation. mAP evaluates detection accuracy by considering both localization precision and recall across various object classes and distances.

    \item \textbf{nuScenes Detection Score (NDS).} NDS provides a comprehensive evaluation by combining mAP with additional detection quality metrics, computed as:
    \begin{equation}
        \label{equ:NDS}
        \text{NDS} = \frac{1}{10}[5 \text{mAP} + \sum_{\text{mTP}\in\text{TP}}(1 - \min(1, \text{mTP}))]~,
    \end{equation}
    where $\text{mTP}$ is the mean true positive metric calculated across all object classes, representing the average number of true positives per class:
    \begin{equation}
        \label{equ:mtp}
        \text{mTP}=\frac{1}{|\mathbb{C}|}\sum_{c\in\mathbb{C}}\text{TP}_{c}~.
    \end{equation}
    NDS captures both detection accuracy and localization precision, making it a more holistic metric for evaluating 3D object detection performance.
\end{itemize}

\section{Additional Quantitative Results}

In this section, we provide a detailed evaluation of class-wise LiDAR semantic segmentation performance, showing the advantages of our approach over existing methods.

\subsection{Class-Wise Linear Probing Results}

\cref{tab_supp:linear_prob} presents the class-wise IoU scores for linear probing experiments. To ensure a fair comparison, we reimplemented ScaLR \cite{puy2024scalr} based on the technical details provided in the original paper and publicly available code. The results show that LiMA consistently outperforms prior state-of-the-art methods across most categories, with particularly notable gains in the segmentation of dynamic objects. This improvement highlights LiMA's capability to capture rich spatiotemporal representations, which are crucial for understanding moving entities in autonomous driving scenarios.

\subsection{Class-Wise Fine-Tuning Results}

We report the class-wise IoU scores for 1\% fine-tuning experiments in \cref{tab_supp:1pct}. Compared to the baseline \cite{puy2024scalr}, LiMA achieves significant performance gains across most categories. These improvements stem from LiMA's ability to leverage temporal feature distillation, effectively capturing long-term dependencies for enhanced feature learning. However, we observe performance degradation in certain underrepresented categories, likely due to class imbalance. This suggests potential avenues for future work, such as incorporating class-aware optimization strategies or re-weighting schemes to mitigate the impact of data sparsity.

\subsection{Knowledge Distillation Strategy}

In the main page, knowledge distillation is employed to transfer long-term temporally fused image features $\mathcal{F}_{d}$ into LiDAR representations $\mathcal{F}_{p}$. In this section, we provide a detailed overview of commonly used distillation strategies and analyze their effectiveness in this context.
\begin{itemize}
    \item \textbf{Cosine Similarity} enforces alignment between feature representations by measuring their angular difference while being invariant to feature magnitudes. The corresponding loss function is formulated as:
    \begin{equation}
        \label{eq:cosine}
        \mathcal{L}_{\text{cos}}(\mathcal{F}_{d}, \mathcal{F}_{p}) = \frac{1}{M} \sum_{i=1}^{M} (1 - \langle \mathbf{f}_{d}^{i}, \mathbf{f}_{p}^{i} \rangle)~,
    \end{equation}
    where $M$ denotes the number of corresponding point-pixel pairs, and $\langle \cdot, \cdot \rangle$ is the dot product. By minimizing $\mathcal{L}_{\text{cos}}$, the LiDAR representation $\mathcal{F}_{p}$ is encouraged to learn features with a similar directional alignment to $\mathcal{F}_{d}$.

    \item \textbf{$\ell_2$ distance} objective minimizes the Euclidean distance between the teacher and student representations, enforcing consistency while preserving feature magnitudes:
    \begin{equation}
        \label{eq:l2}
        \mathcal{L}_{\text{dist}}(\mathcal{F}_{d}, \mathcal{F}_{p}) = \frac{1}{M} \sum_{i=1}^{M} \|\mathbf{f}_{d}^{i} - \mathbf{f}_{p}^{i}\|_2~.
    \end{equation}
    This formulation ensures that $\mathcal{F}_{p}$ closely approximates the fine-grained feature structure of $\mathcal{F}_{d}$.

    \item \textbf{Contrastive Learning} enhances feature discrimination by bringing positive pairs closer while pushing negative pairs apart. The contrastive loss is formulated as:
    \begin{equation}
        \label{eq:contrastive}
        \mathcal{L}_{\text{cont}}(\mathcal{F}_{d}, \mathcal{F}_{p}) = -\frac{1}{M} \sum_{i=1}^{M} \log \frac{e^{\langle \mathbf{f}_{d}^{i}, \mathbf{f}_{p}^{i} \rangle / \tau}}{\sum_{j=1}^{M} e^{\langle \mathbf{f}_{d}^{i}, \mathbf{f}_{p}^{j} \rangle / \tau}}~,
    \end{equation}
    where $\tau > 0$ is a temperature scale factor. This loss function encourages $\mathcal{F}_{p}$ to learn meaningful and discriminative features from $\mathcal{F}_{d}$. In our implementation, we follow PPKT \cite{liu2021ppkt} and randomly sample $4096$ pairs to construct a contrastive objective.

    \item \textbf{Kullback-Leibler (KL) Divergence} measures the discrepancy between two probability distributions, aligning the student’s predictive distribution with the teacher’s. The KL divergence loss is defined as:
    \begin{equation}
        \label{eq:kl}
        \mathcal{L}_{\text{KL}}(\mathcal{F}_{d}, \mathcal{F}_{p}) = \frac{1}{M} \sum_{i=1}^{M} \mathbf{f}_{d}^{i} \log \frac{\mathbf{f}_{d}^{i}}{\mathbf{f}_{p}^{i}}~.
    \end{equation}
    By minimizing $\mathcal{L}_{\text{KL}}$, $\mathcal{F}_{p}$ is encouraged to approximate the predictive distribution of $\mathcal{F}_{d}$, leading to improved generalization performance.
\end{itemize}

\begin{table}[t]
    \centering
    \caption{Comparison of \textbf{different distillation strategies}.}
    \vspace{-0.2cm}
    \label{tab_supp:distillation}
    \resizebox{\linewidth}{!}{
    \begin{tabular}{c|c|cc|c|c}
        \toprule
        \multirow{2}{*}{\textbf{\#}} & \multirow{2}{*}{\textbf{Distillation}} & \multicolumn{2}{c|}{\textbf{nuScenes}} & \textbf{KITTI} & \textbf{Waymo}
        \\
        & & \textbf{LP} & \textbf{1\%} & \textbf{1\%} & \textbf{1\%}
        \\\midrule\midrule
        (a) & Cosine Similarity & $51.23$ & $48.23$ & $47.34$ & $47.84$
        \\
        (b) & $\ell_2$ Distance & \hlgreen{$\mathbf{56.65}$} & \hlgreen{$\mathbf{51.29}$} & \hlgreen{$\mathbf{50.44}$} & \hlgreen{$\mathbf{51.35}$}
        \\
        (c) & Contrastive Learning & $54.23$ & $50.34$ & $48.86$ & $50.23$
        \\
        (d) & KL Divergence & $55.34$ & $49.53$ & $49.43$ & $50.46$
        \\\bottomrule
    \end{tabular}}
\end{table}

\cref{tab_supp:distillation} presents a comparative analysis of various distillation strategies. Our results indicate that the $\ell_2$ distance metric achieves the highest overall performance. The cosine similarity-based approach, which focuses solely on angular alignment, proves inadequate for feature alignment as it disregards magnitude differences -- an essential aspect of representation learning. Contrastive learning aims to enhance feature discrimination by pulling positive pairs closer while pushing negatives apart, but this objective may not be optimal for direct feature matching in distillation. KL divergence effectively aligns predictive distributions but is susceptible to data distribution shifts, particularly in low-data regimes. In contrast, $\ell_2$ distance minimizes Euclidean discrepancy directly, ensuring a more stable and effective optimization objective for knowledge transfer.

\section{Additional Qualitative Results}

In this section, we present additional qualitative examples to provide a visual comparison of the different approaches discussed in the main body of the paper.

\subsection{LiDAR Semantic Segmentation Results}

\begin{figure*}
    \centering
    \includegraphics[width=1.0\linewidth]{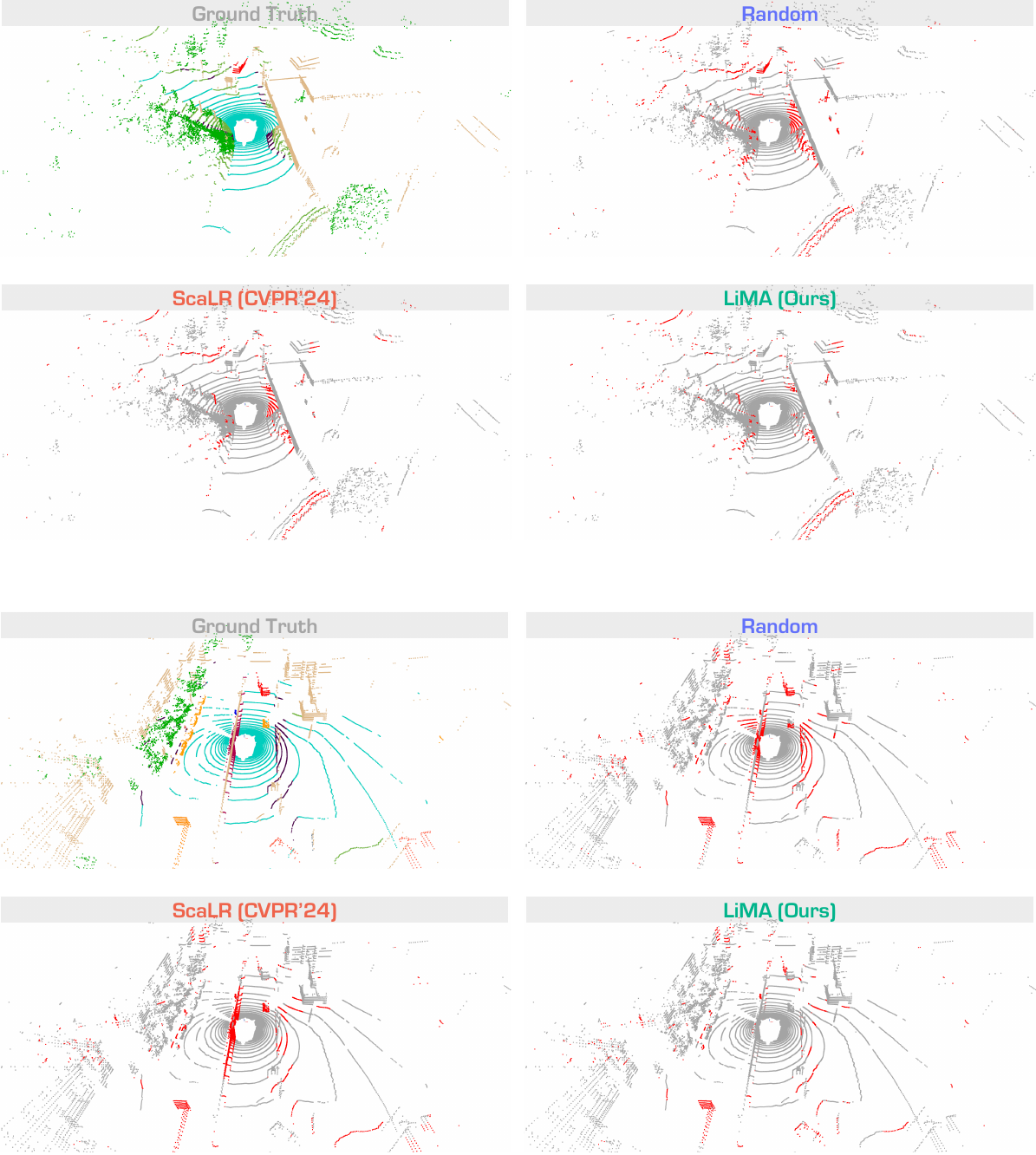}
    \vspace{-0.5cm}
    \caption{\textbf{Qualitative assessments} of state-of-the-art methods, pretrained on \textit{nuScenes} \cite{caesar2020nuscenes} and fine-tuned on \textit{nuScenes} \cite{fong2022panoptic-nuscenes} with $1\%$ annotations. The error maps depict \textcolor{gray}{\textbf{correct}} and \textcolor{term_red}{\textbf{incorrect}} predictions in \textcolor{gray}{\textbf{gray}} and \textcolor{term_red}{\textbf{red}}, respectively. Best viewed in colors.}
    \label{fig_supp:vis_nus}
\end{figure*}

\begin{figure*}
    \centering
    \includegraphics[width=1.0\linewidth]{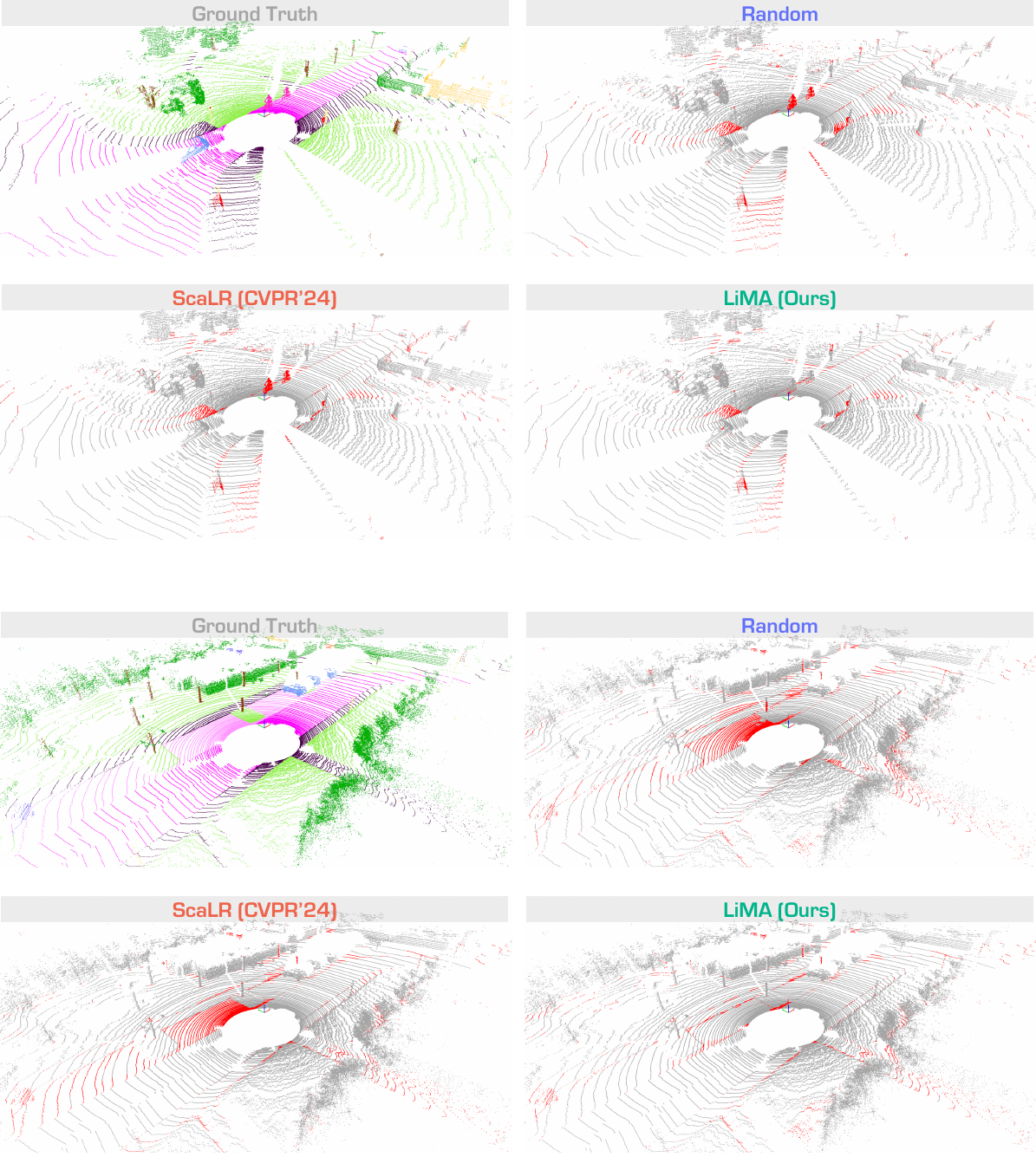}
    \vspace{-0.5cm}
    \caption{\textbf{Qualitative assessments} of state-of-the-art methods, pretrained on \textit{nuScenes} \cite{caesar2020nuscenes} and fine-tuned on \textit{SemanticKITTI} \cite{behley2019semantickitti} with $1\%$ annotations. The error maps depict \textcolor{gray}{\textbf{correct}} and \textcolor{term_red}{\textbf{incorrect}} predictions in \textcolor{gray}{\textbf{gray}} and \textcolor{term_red}{\textbf{red}}, respectively. Best viewed in colors.}
    \label{fig_supp:vis_semkitti}
\end{figure*}

\begin{figure*}
    \centering
    \includegraphics[width=1.0\linewidth]{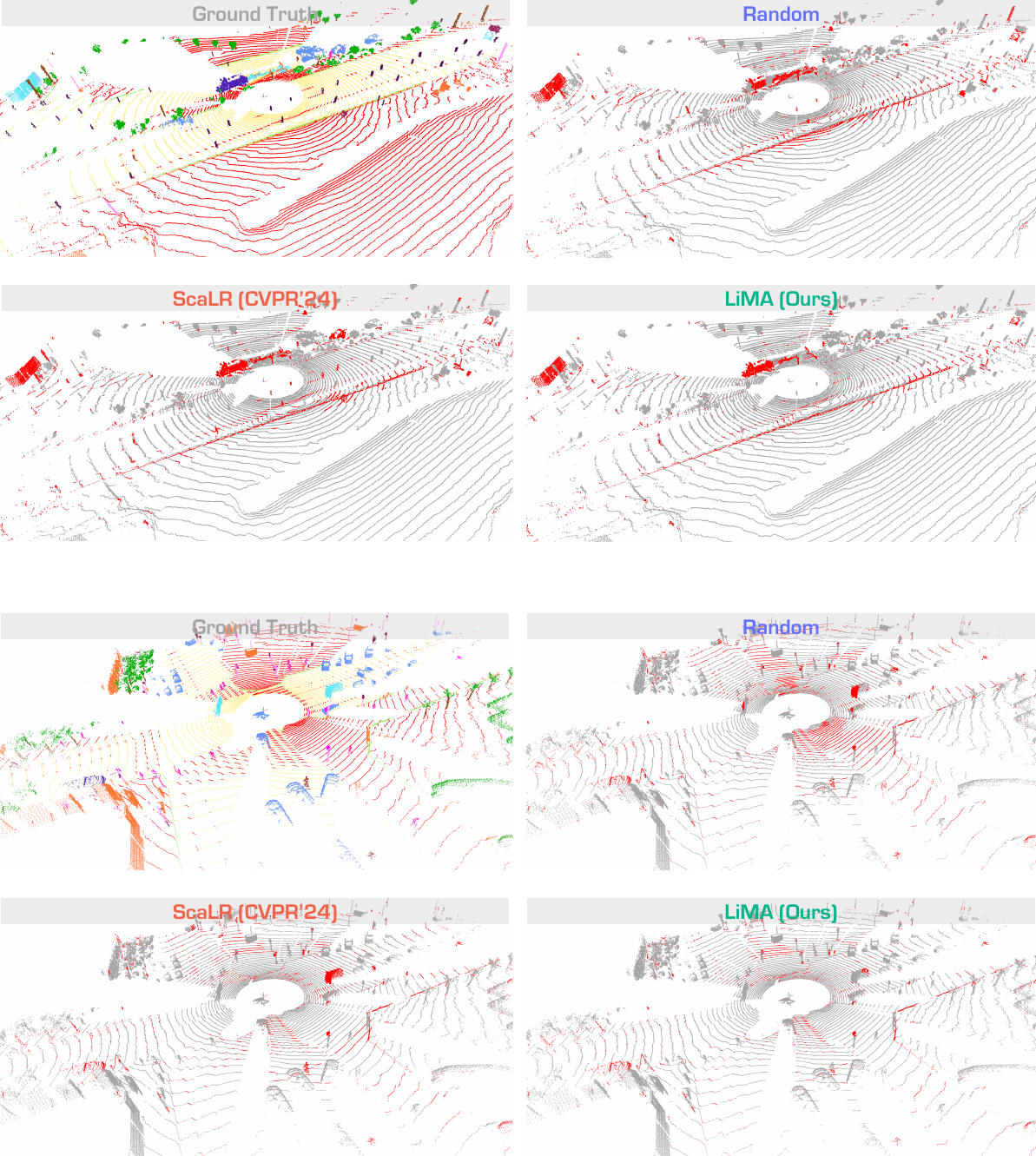}
    \vspace{-0.5cm}
    \caption{\textbf{Qualitative assessments} of state-of-the-art methods, pretrained on \textit{nuScenes} \cite{caesar2020nuscenes} and fine-tuned on \textit{Waymo} \cite{sun2020waymo} with $1\%$ annotations. The error maps depict \textcolor{gray}{\textbf{correct}} and \textcolor{term_red}{\textbf{incorrect}} predictions in \textcolor{gray}{\textbf{gray}} and \textcolor{term_red}{\textbf{red}}, respectively. Best viewed in colors.}
    \label{fig_supp:vis_waymo}
\end{figure*}

\cref{fig_supp:vis_nus}, \cref{fig_supp:vis_semkitti}, and \cref{fig_supp:vis_waymo} showcase qualitative LiDAR semantic segmentation results, comparing LiMA with the baseline method \cite{puy2024scalr} on the \textit{nuScenes} \cite{fong2022panoptic-nuscenes}, \textit{SemanticKITTI} \cite{behley2019semantickitti}, and \textit{Waymo Open} \cite{sun2020waymo} datasets, respectively. The models were pretrained on \textit{nuScenes} \cite{caesar2020nuscenes} and fine-tuned using $1\%$ of the available annotations from each dataset. As shown, LiMA consistently outperforms the baseline in most categories, especially for dynamic objects such as vehicles.

In particular, LiMA can leverage long-term temporal features and align spatial-temporal information, enabling more accurate segmentation in scenarios where objects exhibit rapid motion or change. The error maps further highlight its superior performance, with fewer misclassifications and better localization of the dynamic objects compared to the baseline, showing the robustness of LiMA to both the dataset’s limited annotations and its inherent complexity.

\subsection{3D Object Detection Results}

\begin{figure*}
    \centering
    \includegraphics[width=1.0\linewidth]{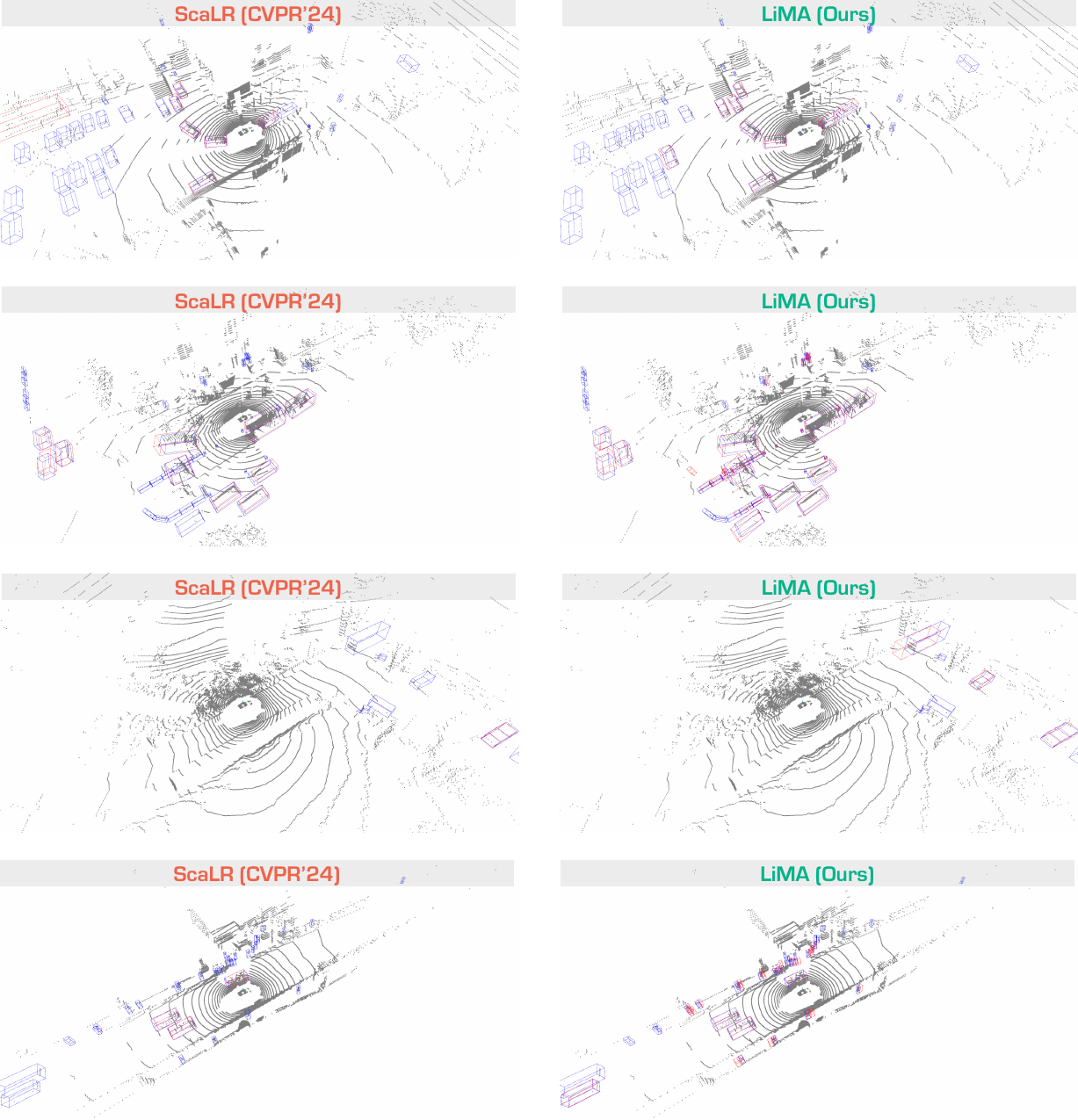}
    \vspace{-0.5cm}
    \caption{\textbf{Qualitative assessments} of object detection, pretrained on \textit{nuScenes} \cite{caesar2020nuscenes} and fine-tuned on \textit{nuScenes} \cite{caesar2020nuscenes} with $5\%$ annotations. The \textcolor{blue}{\textbf{groundtruth}} / \textcolor{red}{\textbf{predicted}} results are highlighted with \textcolor{blue}{\textbf{blue}} / \textcolor{red}{\textbf{red}} boxes, respectively. Best viewed in colors.}
    \label{fig_supp:vis_det}
\end{figure*}

\cref{fig_supp:vis_det} presents qualitative LiDAR detection results, comparing LiMA with the baseline method \cite{puy2024scalr} on the \textit{nuScenes} \cite{caesar2020nuscenes} dataset, where models were fine-tuned using $5\%$ of the available annotations. For better visualization, the confidence threshold is set to $0.5$. As shown, LiMA consistently outperforms the baseline by producing accurate and confident bounding boxes, particularly for small objects.

The long-term memory mechanism effectively propagates and aggregates temporal features, improving object localization and mitigating false positives. In particular, it demonstrates superior performance in detecting dynamic objects such as pedestrians and vehicles, which often suffer from motion-induced distortions. The error analysis highlights LiMA’s ability to maintain spatial-temporal consistency, resulting in fewer missed detections and more stable bounding box predictions. 

\subsection{Cosine Similarity Results}

\begin{figure*}
    \centering
    \includegraphics[width=1.0\linewidth]{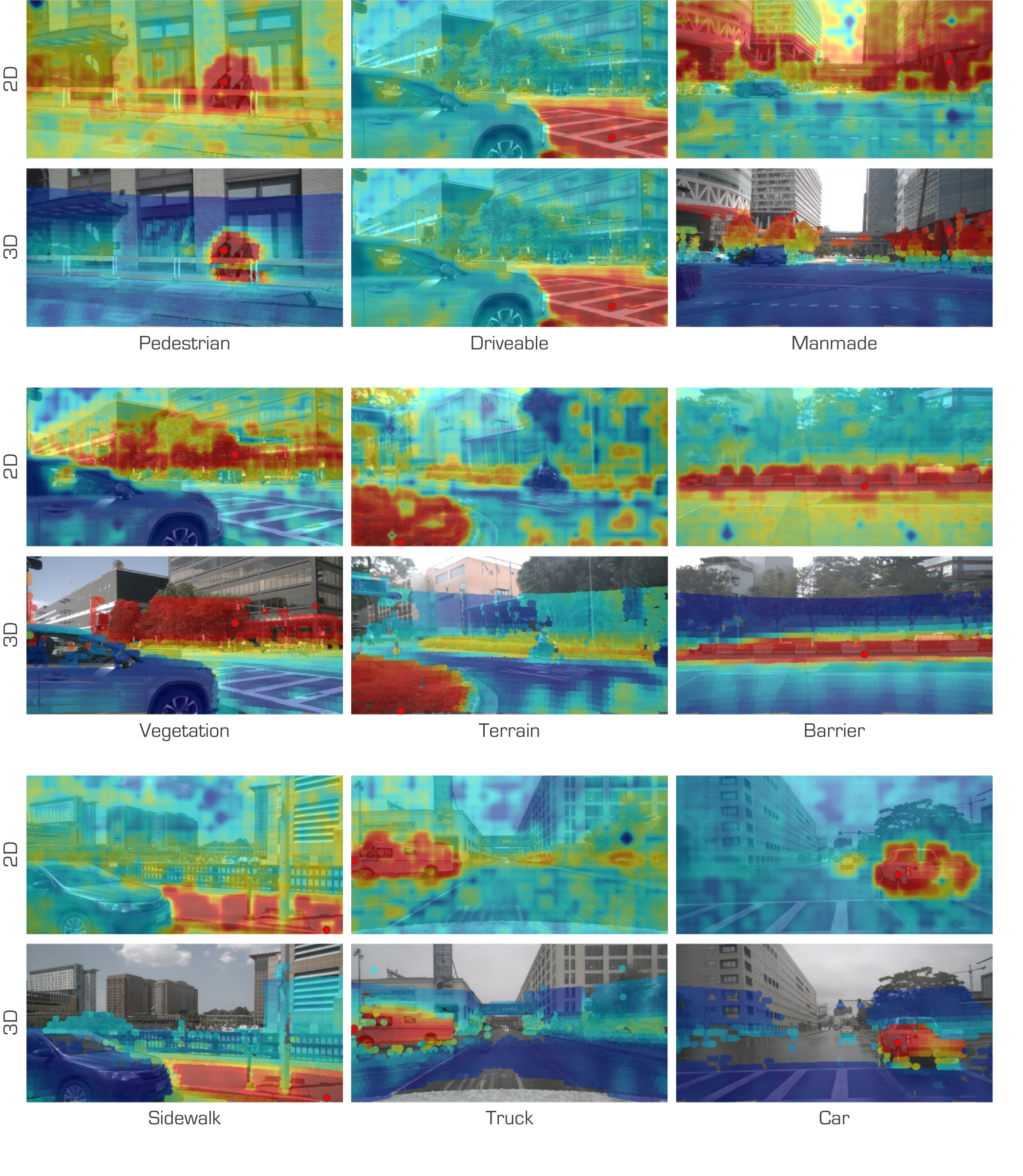}
    \caption{\textbf{Cosine similarity} between a query point (marked as the \textcolor{term_red}{\textbf{red}} dot) and: (1) image features, and (2) LiDAR point features projected onto the image. Colors range from \textcolor{term_red}{\textbf{red}} (indicating \textcolor{term_red}{\textbf{high}} similarity) to \textcolor{term_blue}{\textbf{blue}} (indicating \textcolor{term_blue}{\textbf{low}} similarity). Best viewed in colors.}
    \label{fig_supp:cosine}
\end{figure*}

In \cref{fig_supp:cosine}, we present additional cosine similarity maps computed during the pretraining phase. These maps provide an intuitive understanding of how well LiMA aligns the image and LiDAR point features within the same semantic space. The cosine similarity score, which measures the angular difference between features, is used to evaluate the consistency and semantic relevance between image and LiDAR data.

As depicted, the query point (indicated by the \textcolor{term_red}{red} dot) exhibits high cosine similarity with both the corresponding image and LiDAR point features projected onto the image plane. This demonstrates our ability to effectively bridge the gap between two sensor modalities -- image and LiDAR -- by learning a shared feature space that preserves semantic consistency across them. The resulting high similarity scores (represented in \textcolor{term_red}{red}) indicate that LiMA succeeds in aligning the visual and LiDAR representations, enhancing its capacity to transfer knowledge across modalities and improve feature fusion for downstream tasks.

\section{Broad Impact \& Limitations}

In this section, we discuss the broader impact of our proposed LiMA framework, highlighting its contributions to autonomous perception and beyond. Additionally, we outline potential limitations and areas for future improvement.

\subsection{Broader Impact}

The LiMA framework introduces a novel approach to learning robust LiDAR-based representations through long-term temporal modeling and cross-modal feature alignment. This has several significant implications for both academic research and real-world applications:

\noindent \textbf{Advancing Data-Efficient Perception.} By effectively leveraging pretraining with limited labeled data, LiMA reduces reliance on large-scale human annotations, addressing one of the primary bottlenecks in deep learning for 3D perception. This advancement is particularly valuable for safety-critical applications where data collection is expensive or infeasible, such as autonomous driving and robotics.

\noindent \textbf{Improving Robustness in Dynamic Environments.} Through explicit modeling of temporal and spatial dependencies, LiMA enhances the ability to segment dynamic objects with high accuracy. This contributes to improved situational awareness and decision-making for autonomous systems in complex and rapidly changing environments.

\noindent \textbf{Facilitating Cross-Modal Learning.} The ability to align image and LiDAR features in a shared representation space enhances multi-sensor fusion strategies. This can benefit perception tasks beyond segmentation, including object detection, tracking, and scene understanding, enabling more effective deployment in real-world settings.

\noindent \textbf{Potential for Transfer Learning and Generalization.} The insights from LiMA’s pretraining strategies can be applied to a broader range of 3D vision tasks, fostering new research directions in self-supervised learning, domain adaptation, and transfer learning for sparse and multimodal data.

\subsection{Potential Limitations}

Despite its advantages, LiMA has certain limitations that should be considered for future research:

\noindent \textbf{Sensitivity to Sensor Calibration.} The framework assumes well-calibrated LiDAR and camera sensors for effective cross-modal feature alignment. Misalignment in real-world deployments may lead to suboptimal feature fusion. Future work could explore self-calibration mechanisms or uncertainty-aware fusion techniques.

\noindent \textbf{Dependence on Temporal Consistency.} LiMA relies on long-term temporal information, which may not be optimal in scenarios where past observations are unreliable due to sensor noise, occlusions, or drastic environmental changes. Investigating adaptive temporal modeling techniques could further enhance robustness.

\clearpage
\begin{table*}[t]
    \centering
    \caption{The \textbf{per-class IoU scores} of state-of-the-art pretraining methods pretrained and linear-probed on the \textit{nuScenes} \cite{caesar2020nuscenes,fong2022panoptic-nuscenes} dataset. All scores are given in percentage (\%). The \hlgreen{Best} and \hlred{2nd Best} scores under each group are highlighted in \hlgreen{Green} and \hlred{Red}.}
    \label{tab_supp:linear_prob}
    \resizebox{\linewidth}{!}{
    \begin{tabular}{r|c|cccccccccccccccc}
        \toprule
        \textbf{Method} & \rotatebox{90}{\textbf{mIoU}} & \rotatebox{90}{\textcolor{nu_barrier}{$\blacksquare$}~barrier} & \rotatebox{90}{\textcolor{nu_bicycle}{$\blacksquare$}~bicycle} & \rotatebox{90}{\textcolor{nu_bus}{$\blacksquare$}~bus} & \rotatebox{90}{\textcolor{nu_car}{$\blacksquare$}~car} & \rotatebox{90}{\textcolor{nu_cons}{$\blacksquare$}~construction vehicle~} & \rotatebox{90}{\textcolor{nu_motor}{$\blacksquare$}~motorcycle} & \rotatebox{90}{\textcolor{nu_ped}{$\blacksquare$}~pedestrian} & \rotatebox{90}{\textcolor{nu_cone}{$\blacksquare$}~traffic cone} & \rotatebox{90}{\textcolor{nu_trailer}{$\blacksquare$}~trailer} & \rotatebox{90}{\textcolor{nu_truck}{$\blacksquare$}~truck} & \rotatebox{90}{\textcolor{nu_driv}{$\blacksquare$}~driveable surface} & \rotatebox{90}{\textcolor{nu_flat}{$\blacksquare$}~other flat} & \rotatebox{90}{\textcolor{nu_sidewalk}{$\blacksquare$}~sidewalk} & \rotatebox{90}{\textcolor{nu_terrain}{$\blacksquare$}~terrain} & \rotatebox{90}{\textcolor{nu_manmade}{$\blacksquare$}~manmade} & \rotatebox{90}{\textcolor{nu_veg}{$\blacksquare$}~vegetation}
        \\\midrule\midrule
         \textcolor{gray}{Random} & \textcolor{gray}{$\mathbf{8.1}$} & \textcolor{gray}{$\mathbf{0.5}$} & \textcolor{gray}{$\mathbf{0.0}$} & \textcolor{gray}{$\mathbf{0.0}$} & \textcolor{gray}{$\mathbf{3.9}$} & \textcolor{gray}{$\mathbf{0.0}$} & \textcolor{gray}{$\mathbf{0.0}$} & \textcolor{gray}{$\mathbf{0.0}$} & \textcolor{gray}{$\mathbf{6.4}$} & \textcolor{gray}{$\mathbf{0.0}$} & \textcolor{gray}{$\mathbf{3.9}$} & \textcolor{gray}{$\mathbf{59.6}$} & \textcolor{gray}{$\mathbf{0.0}$} & \textcolor{gray}{$\mathbf{0.1}$} & \textcolor{gray}{$\mathbf{16.2}$} & \textcolor{gray}{$\mathbf{30.6}$} & \textcolor{gray}{$\mathbf{12.0}$}
        \\\midrule
        \multicolumn{18}{l}{\textbf{Distill: None}}
        \\
        PointContrast \cite{xie2020pointcontrast} & \hlred{$\mathbf{21.9}$} & - & - & - & - & - & - & - & - & - & - & - & - & - & - & - & -
        \\
        DepthContrast \cite{zhang2021depthcontrast} & \hlgreen{$\mathbf{22.1}$} & - & - & - & - & - & - & - & - & - & - & - & - & - & - & - & -
        \\
        ALSO \cite{boulch2023also} & - & - & - & - & - & - & - & - & - & - & - & - & - & - & - & - & -
        \\
        BEVContrast \cite{sautier2024bevcontrast} & - & - & - & - & - & - & - & - & - & - & - & - & - & - & - & - & -
        \\\midrule
        \multicolumn{18}{l}{\textbf{Distill: ResNet-50}}
        \\
        PPKT \cite{liu2021ppkt} & $35.9$ & - & - & - & - & - & - & - & - & - & - & - & - & - & - & - & -
        \\
        SLidR \cite{sautier2022slidr} & $39.2$ & \hlred{$\mathbf{44.2}$} & \hlred{$\mathbf{0.0}$} & \hlgreen{$\mathbf{30.8}$} & \hlred{$\mathbf{60.2}$} & \hlred{$\mathbf{15.1}$} & \hlred{$\mathbf{22.4}$} & \hlred{$\mathbf{47.2}$} & \hlred{$\mathbf{27.7}$} & \hlred{$\mathbf{16.3}$} & \hlred{$\mathbf{34.3}$} & \hlred{$\mathbf{80.6}$} & \hlred{$\mathbf{21.8}$} & \hlred{$\mathbf{35.2}$} & \hlred{$\mathbf{48.1}$} & \hlred{$\mathbf{71.0}$} & \hlred{$\mathbf{71.9}$}
        \\
        ST-SLidR \cite{mahmoud2023st-slidr} & $40.5$ & - & - & - & - & - & - & - & - & - & - & - & - & - & - & - & -
        \\
        TriCC \cite{pang2023tricc} & $38.0$ & - & - & - & - & - & - & - & - & - & - & - & - & - & - & - & -
        \\
        Seal \cite{liu2023seal} & \hlred{$\mathbf{45.0}$} & \hlgreen{$\mathbf{54.7}$} & \hlgreen{$\mathbf{5.9}$} & \hlred{$\mathbf{30.6}$} & \hlgreen{$\mathbf{61.7}$} & \hlgreen{$\mathbf{18.9}$} & \hlgreen{$\mathbf{28.8}$} & \hlgreen{$\mathbf{48.1}$} & \hlgreen{$\mathbf{31.0}$} & \hlgreen{$\mathbf{22.1}$} & \hlgreen{$\mathbf{39.5}$} & \hlgreen{$\mathbf{83.8}$} & \hlgreen{$\mathbf{35.4}$} & \hlgreen{$\mathbf{46.7}$} & \hlgreen{$\mathbf{56.9}$} & \hlgreen{$\mathbf{74.7}$} & \hlgreen{$\mathbf{74.7}$}
        \\
        CSC \cite{chen2024csc} & \hlgreen{$\mathbf{46.0}$} & - & - & - & - & - & - & - & - & - & - & - & - & - & - & - & -
        \\
        HVDistill \cite{zhang2024hvdistill} & $39.5$ & - & - & - & - & - & - & - & - & - & - & - & - & - & - & - & -
        \\\midrule
        \multicolumn{18}{l}{\textbf{Distill: ViT-S}}
        \\
        PPKT \cite{liu2021ppkt} & $38.6$ & $43.8$ & $0.0$ & $31.2$ & $53.1$ & $15.2$ & $0.0$ & $42.2$ & $16.5$ & $18.3$ & $33.7$ & $79.1$ & $37.2$ & $45.2$ & $52.7$ & $75.6$ & $74.3$
        \\
        SLidR \cite{sautier2022slidr} & $44.7$ & $45.0$ & $8.2$ & $34.8$ & $58.6$ & \hlred{$\mathbf{23.4}$} & \hlred{$\mathbf{40.2}$} & $43.8$ & $19.0$ & $22.9$ & $40.9$ & $82.7$ & $38.3$ & $47.6$ & $53.9$ & $77.8$ & $77.9$
        \\
        Seal \cite{liu2023seal} & $45.2$ & $48.9$ & \hlgreen{$\mathbf{8.4}$} & $30.7$ & $68.1$ & $17.5$ & $37.7$ & \hlred{$\mathbf{57.7}$} & $17.9$ & $20.9$ & $40.4$ & $83.8$ & $36.6$ & $44.2$ & $54.5$ & $76.2$ & \hlred{$\mathbf{79.3}$}
        \\
        SuperFlow \cite{xu2024superflow} & $46.4$ & $49.8$ & \hlred{$\mathbf{6.8}$} & $45.9$ & $63.4$ & $18.5$ & $31.0$ & \hlgreen{$\mathbf{60.3}$} & \hlred{$\mathbf{28.1}$} & \hlred{$\mathbf{25.4}$} & \hlred{$\mathbf{47.4}$} & $86.2$ & $38.4$ & $47.4$ & $56.7$ & $74.9$ & $77.8$
        \\
        ScaLR \cite{puy2024scalr} & \hlred{$\mathbf{49.7}$} & \hlred{$\mathbf{58.5}$} & $3.2$ & \hlred{$\mathbf{62.4}$} & \hlred{$\mathbf{68.8}$} & $20.2$ & $32.3$ & $49.0$ & \hlgreen{$\mathbf{31.8}$} & $21.7$ & $45.9$ & \hlred{$\mathbf{90.0}$} & \hlred{$\mathbf{39.5}$} & \hlred{$\mathbf{53.1}$} & \hlred{$\mathbf{62.1}$} & \hlred{$\mathbf{78.0}$} & $78.1$
        \\
        \textbf{LiMA} & \hlgreen{$\mathbf{54.8}$} & \hlgreen{$\mathbf{61.9}$} & $3.5$ & \hlgreen{$\mathbf{71.6}$} & \hlgreen{$\mathbf{73.3}$} & \hlgreen{$\mathbf{29.3}$} & \hlgreen{$\mathbf{46.8}$} & $53.9$ & \hlgreen{$\mathbf{31.8}$} & \hlgreen{$\mathbf{27.7}$} & \hlgreen{$\mathbf{55.5}$} & \hlgreen{$\mathbf{91.9}$} & \hlgreen{$\mathbf{43.5}$} & \hlgreen{$\mathbf{59.7}$} & \hlgreen{$\mathbf{65.8}$} & \hlgreen{$\mathbf{80.2}$} & \hlgreen{$\mathbf{80.0}$}
        \\\midrule
        \multicolumn{18}{l}{\textbf{Distill: ViT-B}}
        \\
        PPKT \cite{liu2021ppkt} & $40.0$ & $29.6$ & $0.0$ & $30.7$ & $55.8$ & $6.3$ & $22.4$ & $56.7$ & $18.1$ & \hlgreen{$\mathbf{24.3}$} & $42.7$ & $82.3$ & $33.2$ & $45.1$ & $53.4$ & $71.3$ & $75.7$
        \\
        SLidR \cite{sautier2022slidr} & $45.4$ & $46.7$ & $7.8$ & $46.5$ & $58.7$ & $23.9$ & $34.0$ & $47.8$ & $17.1$ & \hlred{$\mathbf{23.7}$} & $41.7$ & $83.4$ & $39.4$ & $47.0$ & $54.6$ & $76.6$ & $77.8$
        \\
        Seal \cite{liu2023seal} & $46.6$ & $49.3$ & \hlred{$\mathbf{8.2}$} & $35.1$ & $70.8$ & $22.1$ & $41.7$ & \hlred{$\mathbf{57.4}$} & $15.2$ & $21.6$ & $42.6$ & $84.5$ & $38.1$ & $46.8$ & $55.4$ & $77.2$ & \hlred{$\mathbf{79.5}$}
        \\
        SuperFlow \cite{xu2024superflow} & $47.7$ & $45.8$ & \hlgreen{$\mathbf{12.4}$} & $52.6$ & $67.9$ & $17.2$ & \hlred{$\mathbf{40.8}$} & \hlgreen{$\mathbf{59.5}$} & $25.4$ & $21.0$ & $47.6$ & $85.8$ & $37.2$ & $48.4$ & $56.6$ & $76.2$ & $78.2$
        \\
        ScaLR \cite{puy2024scalr} & \hlred{$\mathbf{51.9}$} & \hlred{$\mathbf{61.6}$} & $3.1$ & \hlred{$\mathbf{70.2}$} & \hlred{$\mathbf{70.9}$} & \hlred{$\mathbf{25.2}$} & $29.5$ & $48.3$ & \hlred{$\mathbf{32.8}$} & $22.3$ & \hlred{$\mathbf{49.9}$} & \hlred{$\mathbf{90.9}$} & \hlred{$\mathbf{45.3}$} & \hlred{$\mathbf{57.9}$} & \hlred{$\mathbf{64.9}$} & \hlred{$\mathbf{79.2}$} & $78.3$
        \\
        \textbf{LiMA} & \hlgreen{$\mathbf{56.7}$} & \hlgreen{$\mathbf{63.4}$} & $4.1$ & \hlgreen{$\mathbf{73.7}$} & \hlgreen{$\mathbf{76.4}$} & \hlgreen{$\mathbf{32.8}$} & \hlgreen{$\mathbf{43.4}$} & $54.6$ & \hlgreen{$\mathbf{38.8}$} & \hlgreen{$\mathbf{24.3}$} & \hlgreen{$\mathbf{57.2}$} & \hlgreen{$\mathbf{92.9}$} & \hlgreen{$\mathbf{51.3}$} & \hlgreen{$\mathbf{63.8}$} & \hlgreen{$\mathbf{68.4}$} & \hlgreen{$\mathbf{81.0}$} & \hlgreen{$\mathbf{80.5}$}
        \\\midrule
        \multicolumn{18}{l}{\textbf{Distill: ViT-L}}
        \\
        PPKT \cite{liu2021ppkt} & $41.6$ & $30.5$ & $0.0$ & $32.0$ & $57.3$ & $8.7$ & $24.0$ & \hlred{$\mathbf{58.1}$} & $19.5$ & \hlred{$\mathbf{24.9}$} & $44.1$ & $83.1$ & $34.5$ & $45.9$ & $55.4$ & $72.5$ & $76.4$
        \\
        SLidR \cite{sautier2022slidr} & $45.7$ & $46.9$ & \hlred{$\mathbf{6.9}$} & $44.9$ & $60.8$ & $22.7$ & $40.6$ & $44.7$ & $17.4$ & $23.0$ & $40.4$ & $83.6$ & $39.9$ & $47.8$ & $55.2$ & $78.1$ & $78.3$
        \\
        Seal \cite{liu2023seal} & $46.8$ & $53.1$ & \hlred{$\mathbf{6.9}$} & $35.0$ & $65.0$ & $22.0$ & \hlred{$\mathbf{46.1}$} & \hlgreen{$\mathbf{59.2}$} & $16.2$ & $23.0$ & $41.8$ & $84.7$ & $35.8$ & $46.6$ & $55.5$ & $78.4$ & \hlred{$\mathbf{79.8}$}
        \\
        SuperFlow \cite{xu2024superflow} & $48.0$ & $52.3$ & \hlgreen{$\mathbf{12.7}$} & $46.5$ & $64.7$ & $21.4$ & $44.9$ & $56.2$ & $26.7$ & $19.9$ & $43.2$ & $84.2$ & $38.1$ & $47.4$ & $56.9$ & $76.0$ & $79.2$
        \\
        ScaLR \cite{puy2024scalr} & \hlred{$\mathbf{51.8}$} & \hlred{$\mathbf{61.2}$} & $3.4$ & \hlred{$\mathbf{65.4}$} & \hlred{$\mathbf{72.4}$} & \hlred{$\mathbf{25.6}$} & $34.3$ & $51.7$ & \hlred{$\mathbf{28.8}$} & $23.6$ & \hlred{$\mathbf{50.4}$} & \hlred{$\mathbf{90.6}$} & \hlred{$\mathbf{44.1}$} & \hlred{$\mathbf{55.6}$} & \hlred{$\mathbf{64.5}$} & \hlred{$\mathbf{79.3}$} & $79.5$
        \\
        \textbf{LiMA} & \hlgreen{$\mathbf{56.7}$} & \hlgreen{$\mathbf{63.6}$} & $3.5$ & \hlgreen{$\mathbf{72.4}$} & \hlgreen{$\mathbf{75.0}$} & \hlgreen{$\mathbf{33.7}$} & \hlgreen{$\mathbf{48.5}$} & $55.7$ & \hlgreen{$\mathbf{37.4}$} & \hlgreen{$\mathbf{25.3}$} & \hlgreen{$\mathbf{59.0}$} & \hlgreen{$\mathbf{92.6}$} & \hlgreen{$\mathbf{48.3}$} & \hlgreen{$\mathbf{62.0}$} & \hlgreen{$\mathbf{68.1}$} & \hlgreen{$\mathbf{81.1}$} & \hlgreen{$\mathbf{80.6}$}
        \\\bottomrule
    \end{tabular}}
\end{table*}

\clearpage
\begin{table*}[t]
    \centering
    \caption{The \textbf{per-class IoU scores} of state-of-the-art pretraining methods pretrained and fine-tuned on \textit{nuScenes} \cite{caesar2020nuscenes,fong2022panoptic-nuscenes} dataset with $1\%$ annotations. All scores are given in percentage (\%). The \hlgreen{Best} and \hlred{2nd Best} scores under each group are highlighted in \hlgreen{Green} and \hlred{Red}.}
    \label{tab_supp:1pct}
    \resizebox{\linewidth}{!}{
    \begin{tabular}{r|c|cccccccccccccccc}
        \toprule
        \textbf{Method} & \rotatebox{90}{\textbf{mIoU}} & \rotatebox{90}{\textcolor{nu_barrier}{$\blacksquare$}~barrier} & \rotatebox{90}{\textcolor{nu_bicycle}{$\blacksquare$}~bicycle} & \rotatebox{90}{\textcolor{nu_bus}{$\blacksquare$}~bus} & \rotatebox{90}{\textcolor{nu_car}{$\blacksquare$}~car} & \rotatebox{90}{\textcolor{nu_cons}{$\blacksquare$}~construction vehicle~} & \rotatebox{90}{\textcolor{nu_motor}{$\blacksquare$}~motorcycle} & \rotatebox{90}{\textcolor{nu_ped}{$\blacksquare$}~pedestrian} & \rotatebox{90}{\textcolor{nu_cone}{$\blacksquare$}~traffic cone} & \rotatebox{90}{\textcolor{nu_trailer}{$\blacksquare$}~trailer} & \rotatebox{90}{\textcolor{nu_truck}{$\blacksquare$}~truck} & \rotatebox{90}{\textcolor{nu_driv}{$\blacksquare$}~driveable surface} & \rotatebox{90}{\textcolor{nu_flat}{$\blacksquare$}~other flat} & \rotatebox{90}{\textcolor{nu_sidewalk}{$\blacksquare$}~sidewalk} & \rotatebox{90}{\textcolor{nu_terrain}{$\blacksquare$}~terrain} & \rotatebox{90}{\textcolor{nu_manmade}{$\blacksquare$}~manmade} & \rotatebox{90}{\textcolor{nu_veg}{$\blacksquare$}~vegetation}
        \\\midrule\midrule
        \textcolor{gray}{Random} & \textcolor{gray}{$\mathbf{30.3}$} & \textcolor{gray}{$\mathbf{0.0}$} & \textcolor{gray}{$\mathbf{0.0}$} & \textcolor{gray}{$\mathbf{8.1}$} & \textcolor{gray}{$\mathbf{65.0}$} & \textcolor{gray}{$\mathbf{0.1}$} & \textcolor{gray}{$\mathbf{6.6}$} & \textcolor{gray}{$\mathbf{21.0}$} & \textcolor{gray}{$\mathbf{9.0}$} & \textcolor{gray}{$\mathbf{9.3}$} & \textcolor{gray}{$\mathbf{25.8}$} & \textcolor{gray}{$\mathbf{89.5}$} & \textcolor{gray}{$\mathbf{14.8}$} & \textcolor{gray}{$\mathbf{41.7}$} & \textcolor{gray}{$\mathbf{48.7}$} & \textcolor{gray}{$\mathbf{72.4}$} & \textcolor{gray}{$\mathbf{73.3}$}
        \\\midrule
        \multicolumn{18}{l}{\textbf{Distill: None}}
        \\
        PointContrast \cite{xie2020pointcontrast} & $32.5$ & \hlgreen{$\mathbf{0.0}$} & \hlred{$\mathbf{1.0}$} & $5.6$ & \hlred{$\mathbf{67.4}$} & $0.0$ & \hlred{$\mathbf{3.3}$} & \hlred{$\mathbf{31.6}$} & $5.6$ & \hlred{$\mathbf{12.1}$} & \hlred{$\mathbf{30.8}$} & \hlgreen{$\mathbf{91.7}$} & \hlred{$\mathbf{21.9}$} & \hlgreen{$\mathbf{48.4}$} & \hlred{$\mathbf{50.8}$} & \hlred{$\mathbf{75.0}$} & \hlred{$\mathbf{74.6}$}
        \\
        DepthContrast \cite{zhang2021depthcontrast} & $31.7$ & \hlgreen{$\mathbf{0.0}$} & $0.6$ & \hlred{$\mathbf{6.5}$} & $64.7$ & \hlred{$\mathbf{0.2}$} & \hlgreen{$\mathbf{5.1}$} & $29.0$ & \hlgreen{$\mathbf{9.5}$} & \hlred{$\mathbf{12.1}$} & $29.9$ & \hlred{$\mathbf{90.3}$} & $17.8$ & \hlred{$\mathbf{44.4}$} & $49.5$ & $73.5$ & $74.0$
        \\
        ALSO \cite{boulch2023also} & \hlred{$37.7$} & - & - & - & - & - & - & - & - & - & - & - & - & - & - & - & -
        \\
        BEVContrast \cite{sautier2024bevcontrast} & \hlgreen{$\mathbf{37.9}$} & \hlgreen{$\mathbf{0.0}$} & \hlgreen{$\mathbf{1.3}$} & \hlgreen{$\mathbf{32.6}$} & \hlgreen{$\mathbf{74.3}$} & \hlgreen{$\mathbf{1.1}$} & $0.9$ & \hlgreen{$\mathbf{41.3}$} & \hlred{$\mathbf{8.1}$} & \hlgreen{$\mathbf{24.1}$} & \hlgreen{$\mathbf{40.9}$} & $89.8$ & \hlgreen{$\mathbf{36.2}$} & $44.0$ & \hlgreen{$\mathbf{52.1}$} & \hlgreen{$\mathbf{79.9}$} & \hlgreen{$\mathbf{79.7}$}
        \\\midrule
        \multicolumn{18}{l}{\textbf{Distill: ResNet-50}}
        \\
        PPKT \cite{liu2021ppkt} & $37.8$ & \hlgreen{$\mathbf{0.0}$} & \hlred{$\mathbf{2.2}$} & $20.7$ & \hlred{$\mathbf{75.4}$} & $1.2$ & $13.2$ & $45.6$ & $8.5$ & $17.5$ & $38.4$ & \hlred{$\mathbf{92.5}$} & $19.2$ & $52.3$ & $56.8$ & $80.1$ & $80.9$
        \\
        SLidR \cite{sautier2022slidr} & $38.8$ & \hlgreen{$\mathbf{0.0}$} & $1.8$ & $15.4$ & $73.1$ & \hlred{$\mathbf{1.9}$} & $19.9$ & $47.2$ & $17.1$ & $14.5$ & $34.5$ & $92.0$ & $27.1$ & $53.6$ & \hlred{$\mathbf{61.0}$} & $79.8$ & $82.3$
        \\
        ST-SLidR \cite{mahmoud2023st-slidr} & $40.8$ & - & - & - & - & - & - & - & - & - & - & - & - & - & - & - & -
        \\
        TriCC \cite{pang2023tricc} & $41.2$ & - & - & - & - & - & - & - & - & - & - & - & - & - & - & - & -
        \\
        Seal \cite{liu2023seal} & \hlred{$\mathbf{45.8}$} & \hlgreen{$\mathbf{0.0}$} & \hlgreen{$\mathbf{9.4}$} & \hlred{$\mathbf{32.6}$} & \hlgreen{$\mathbf{77.5}$} & \hlgreen{$\mathbf{10.4}$} & \hlred{$\mathbf{28.0}$} & \hlred{$\mathbf{53.0}$} & \hlred{$\mathbf{25.0}$} & \hlgreen{$\mathbf{30.9}$} & \hlgreen{$\mathbf{49.7}$} & \hlgreen{$\mathbf{94.0}$} & \hlgreen{$\mathbf{33.7}$} & \hlgreen{$\mathbf{60.1}$} & $59.6$ & \hlgreen{$\mathbf{83.9}$} & \hlred{$\mathbf{83.4}$}
        \\
        CSC \cite{chen2024csc} & \hlgreen{$\mathbf{47.0}$} & \hlgreen{$\mathbf{0.0}$} & $0.0$ & \hlgreen{$\mathbf{58.7}$} & $74.0$ & $0.1$ & \hlgreen{$\mathbf{40.9}$} & \hlgreen{$\mathbf{58.9}$} & \hlgreen{$\mathbf{31.8}$} & \hlred{$\mathbf{23.7}$} & \hlred{$\mathbf{45.1}$} & \hlred{$\mathbf{92.5}$} & \hlred{$\mathbf{33.0}$} & \hlred{$\mathbf{56.4}$} & \hlgreen{$\mathbf{62.4}$} & \hlred{$\mathbf{81.6}$} & \hlgreen{$\mathbf{84.2}$}
        \\
        HVDistill \cite{zhang2024hvdistill} & $42.7$ & - & - & - & - & - & - & - & - & - & - & - & - & - & - & - & -
        \\\midrule
        \multicolumn{18}{l}{\textbf{Distill: ViT-S}}
        \\
        PPKT \cite{liu2021ppkt} & $40.6$ & $0.0$ & $0.0$ & $25.2$ & $73.5$ & $9.1$ & $6.9$ & $51.4$ & $8.6$ & $11.3$ & $31.1$ & $93.2$ & \hlred{$\mathbf{41.7}$} & $58.3$ & \hlred{$\mathbf{64.0}$} & $82.0$ & \hlred{$\mathbf{82.6}$}
        \\
        SLidR \cite{sautier2022slidr} & $41.2$ & $0.0$ & $0.0$ & $26.6$ & $72.0$ & \hlred{$\mathbf{12.4}$} & $15.8$ & $51.4$ & \hlred{$\mathbf{22.9}$} & $11.7$ & $35.3$ & $92.9$ & $36.3$ & $58.7$ & $63.6$ & $81.2$ & $82.3$
        \\
        Seal \cite{liu2023seal} & $44.3$ & $20.0$ & $0.0$ & $19.4$ & \hlred{$\mathbf{74.7}$} & $10.6$ & \hlred{$\mathbf{45.7}$} & \hlgreen{$\mathbf{60.3}$} & \hlgreen{$\mathbf{29.2}$} & \hlred{$\mathbf{17.4}$} & $38.1$ & $93.2$ & $26.0$ & \hlred{$\mathbf{58.8}$} & \hlgreen{$\mathbf{64.5}$} & $81.9$ & $81.9$
        \\
        SuperFlow \cite{xu2024superflow} & \hlred{$\mathbf{47.8}$} & \hlred{$\mathbf{38.2}$} & \hlred{$\mathbf{1.8}$} & $25.8$ & \hlgreen{$\mathbf{79.0}$} & \hlgreen{$\mathbf{15.3}$} & $43.6$ & \hlgreen{$\mathbf{60.3}$} & $0.0$ & \hlgreen{$\mathbf{28.4}$} & \hlred{$\mathbf{55.4}$} & \hlred{$\mathbf{93.7}$} & $28.8$ & \hlgreen{$\mathbf{59.1}$} & $59.9$ & \hlgreen{$\mathbf{83.5}$} & \hlgreen{$\mathbf{83.1}$}
        \\
        ScaLR \cite{puy2024scalr} & $45.9$ & $35.8$ & \hlgreen{$\mathbf{6.0}$} & \hlred{$\mathbf{57.0}$} & $72.7$ & $0.6$ & $42.7$ & $47.5$ & $3.7$ & $8.4$ & $55.3$ & $92.7$ & $28.3$ & $56.3$ & $62.7$ & \hlred{$\mathbf{83.3}$} & $81.3$
        \\
        \textbf{LiMA} & \hlgreen{$\mathbf{48.8}$} & \hlgreen{$\mathbf{42.2}$} & $0.8$ & \hlgreen{$\mathbf{66.0}$} & $74.4$ & $0.0$ & \hlgreen{$\mathbf{47.3}$} & \hlred{$\mathbf{54.2}$} & $0.0$ & $14.8$ & \hlgreen{$\mathbf{59.2}$} & \hlgreen{$\mathbf{93.8}$} & \hlgreen{$\mathbf{42.6}$} & $58.6$ & $62.0$ & $83.0$ & $81.1$
        \\\midrule
        \multicolumn{18}{l}{\textbf{Distill: ViT-B}}
        \\
        PPKT \cite{liu2021ppkt} & $40.9$ & $0.0$ & $0.0$ & $24.5$ & $73.5$ & \hlgreen{$\mathbf{12.2}$} & $7.0$ & $51.0$ & \hlred{$\mathbf{13.5}$} & $15.4$ & $36.3$ & $93.1$ & $40.4$ & $59.2$ & $63.5$ & $81.7$ & $82.2$
        \\
        SLidR \cite{sautier2022slidr} & $41.6$ & $0.0$ & $0.0$ & $26.7$ & $73.4$ & $10.3$ & $16.9$ & $51.3$ & \hlgreen{$\mathbf{23.3}$} & $12.7$ & $38.1$ & $93.0$ & $37.7$ & $58.8$ & $63.4$ & $81.6$ & \hlred{$\mathbf{82.7}$}
        \\
        Seal \cite{liu2023seal} & $46.0$ & $43.0$ & $0.0$ & $26.7$ & \hlgreen{$\mathbf{81.3}$} & $9.9$ & $41.3$ & \hlred{$\mathbf{56.2}$} & $0.0$ & \hlred{$\mathbf{21.7}$} & $51.6$ & $93.6$ & \hlgreen{$\mathbf{42.3}$} & \hlgreen{$\mathbf{62.8}$} & $64.7$ & $82.6$ & \hlred{$\mathbf{82.7}$}
        \\
        SuperFlow \cite{xu2024superflow} & $48.1$ & $39.1$ & $0.9$ & $30.0$ & \hlred{$\mathbf{80.7}$} & $10.3$ & \hlgreen{$\mathbf{47.1}$} & \hlgreen{$\mathbf{59.5}$} & $5.1$ & \hlgreen{$\mathbf{27.6}$} & $55.4$ & $93.7$ & $29.1$ & $61.1$ & $63.5$ & $82.7$ & \hlgreen{$\mathbf{83.6}$}
        \\
        ScaLR \cite{puy2024scalr} & \hlred{$\mathbf{48.9}$} & \hlred{$\mathbf{52.8}$} & \hlgreen{$\mathbf{4.1}$} & \hlred{$\mathbf{66.6}$} & $71.7$ & $0.2$ & \hlred{$\mathbf{44.0}$} & $46.5$ & $11.1$ & $5.8$ & \hlred{$\mathbf{56.1}$} & \hlred{$\mathbf{93.8}$} & $35.8$ & \hlred{$\mathbf{61.7}$} & \hlgreen{$\mathbf{66.8}$} & \hlred{$\mathbf{83.7}$} & $81.8$
        \\
        \textbf{LiMA} & \hlgreen{$\mathbf{51.3}$} & \hlgreen{$\mathbf{53.2}$} & \hlred{$\mathbf{3.6}$} & \hlgreen{$\mathbf{69.0}$} & $78.1$ & \hlred{$\mathbf{11.0}$} & \hlgreen{$\mathbf{47.1}$} & $52.4$ & $7.5$ & $4.9$ & \hlgreen{$\mathbf{62.2}$} & \hlgreen{$\mathbf{94.0}$} & \hlred{$\mathbf{40.5}$} & $60.3$ & \hlred{$\mathbf{66.0}$} & \hlgreen{$\mathbf{85.1}$} & $82.6$
        \\\midrule
        \multicolumn{18}{l}{\textbf{Distill: ViT-L}}
        \\
        PPKT \cite{liu2021ppkt} & $42.1$ & $0.0$ & $0.0$ & $24.4$ & $78.8$ & $15.1$ & $9.2$ & $54.2$ & $14.3$ & $12.9$ & $39.1$ & $92.9$ & $37.8$ & $59.8$ & $64.9$ & $82.3$ & $83.6$
        \\
        SLidR \cite{sautier2022slidr} & $42.8$ & $0.0$ & $0.0$ & $23.9$ & $78.8$ & $15.2$ & $20.9$ & $55.0$ & \hlgreen{$\mathbf{28.0}$} & $17.4$ & $41.4$ & $92.2$ & $41.2$ & $58.0$ & $64.0$ & $81.8$ & $82.7$
        \\
        Seal \cite{liu2023seal} & $46.3$ & $41.8$ & $0.0$ & $23.8$ & \hlgreen{$\mathbf{81.4}$} & \hlgreen{$\mathbf{17.7}$} & $46.3$ & \hlred{$\mathbf{58.6}$} & $0.0$ & \hlred{$\mathbf{23.4}$} & $54.7$ & $93.8$ & \hlred{$\mathbf{41.4}$} & \hlred{$\mathbf{62.5}$} & $65.0$ & $83.8$ & \hlred{$\mathbf{83.8}$}
        \\
        SuperFlow \cite{xu2024superflow} & \hlred{$\mathbf{50.0}$} & $44.5$ & $0.9$ & $22.4$ & \hlred{$\mathbf{80.8}$} & \hlred{$\mathbf{17.1}$} & \hlred{$\mathbf{50.2}$} & \hlgreen{$\mathbf{60.9}$} & \hlred{$\mathbf{21.0}$} & \hlgreen{$\mathbf{25.1}$} & $55.1$ & \hlred{$\mathbf{93.9}$} & $35.8$ & $61.5$ & $62.6$ & $83.7$ & $83.7$
        \\
        ScaLR \cite{puy2024scalr} & $49.1$ & \hlred{$\mathbf{46.5}$} & \hlred{$\mathbf{4.9}$} & \hlred{$\mathbf{70.5}$} & $77.0$ & $2.5$ & $45.9$ & $47.7$ & $9.1$ & $4.9$ & \hlred{$\mathbf{55.6}$} & $93.8$ & $35.4$ & $59.4$ & \hlred{$\mathbf{66.2}$} & \hlred{$\mathbf{84.1}$} & $82.5$
        \\
        \textbf{LiMA} & \hlgreen{$\mathbf{53.2}$} & \hlgreen{$\mathbf{54.0}$} & \hlgreen{$\mathbf{5.5}$} & \hlgreen{$\mathbf{71.3}$} & $76.7$ & $11.2$ & \hlgreen{$\mathbf{59.3}$} & $54.2$ & $10.2$ & $9.4$ & \hlgreen{$\mathbf{61.0}$} & \hlgreen{$\mathbf{94.7}$} & \hlgreen{$\mathbf{43.4}$} & \hlgreen{$\mathbf{63.4}$} & \hlgreen{$\mathbf{68.9}$} & \hlgreen{$\mathbf{84.2}$} & \hlgreen{$\mathbf{84.1}$}
        \\\bottomrule
    \end{tabular}}
\end{table*}

\clearpage

\clearpage

\clearpage

\clearpage

\clearpage

\clearpage\clearpage
\section{Public Resources Used}
\label{sec:supp_acknowledge}

In this section, we acknowledge the use of the following public resources, during the course of this work.

\subsection{Public Codebase Used}

We acknowledge the use of the following public codebase, during the course of this work:

\begin{itemize}
    \item MMEngine\footnote{\url{https://github.com/open-mmlab/mmengine}.} \dotfill Apache License 2.0
    \item MMCV\footnote{\url{https://github.com/open-mmlab/mmcv}.} \dotfill Apache License 2.0
    \item MMDetection\footnote{\url{https://github.com/open-mmlab/mmdetection}.} \dotfill Apache License 2.0
    \item MMDetection3D\footnote{\url{https://github.com/open-mmlab/mmdetection3d}.} \dotfill Apache License 2.0
    \item OpenPCSeg\footnote{\url{https://github.com/PJLab-ADG/OpenPCSeg}.} \dotfill Apache License 2.0
\end{itemize}

\subsection{Public Datasets Used}

We acknowledge the use of the following public datasets, during the course of this work:

\begin{itemize}
    \item nuScenes\footnote{\url{https://www.nuscenes.org/nuscenes}.} \dotfill CC BY-NC-SA 4.0
    \item SemanticKITTI\footnote{\url{http://semantic-kitti.org}.} \dotfill CC BY-NC-SA 4.0
    \item Waymo Open\footnote{\url{https://waymo.com/open}.} \dotfill Waymo Dataset License
    \item ScribbleKITTI\footnote{\url{https://github.com/ouenal/scribblekitti}.} \dotfill Unknown
    \item RELLIS-3D\footnote{\url{https://github.com/unmannedlab/RELLIS-3D}.} \dotfill CC BY-NC-SA 3.0
    \item SemanticPOSS\footnote{\url{http://www.poss.pku.edu.cn/semanticposs.html}.} \dotfill CC BY-NC-SA 3.0
    \item SemanticSTF\footnote{\url{https://github.com/xiaoaoran/SemanticSTF}.} \dotfill CC BY-NC-SA 4.0
    \item SynLiDAR\footnote{\url{https://github.com/xiaoaoran/SynLiDAR}.} \dotfill MIT License
    \item DAPS-3D\footnote{\url{https://github.com/subake/DAPS3D}.} \dotfill MIT License
    \item Synth4D\footnote{\url{https://github.com/saltoricristiano/gipso-sfouda}.} \dotfill GPL-3.0 License
    \item Robo3D\footnote{\url{https://github.com/ldkong1205/Robo3D}.} \dotfill CC BY-NC-SA 4.0
\end{itemize}

\subsection{Public Implementations Used}

We acknowledge the use of the following implementations, during the course of this work:

\begin{itemize}
    \item nuscenes-devkit\footnote{\url{https://github.com/nutonomy/nuscenes-devkit}.} \dotfill Apache License 2.0
    \item semantic-kitti-api\footnote{\url{https://github.com/PRBonn/semantic-kitti-api}.} \dotfill MIT License
    \item waymo-open-dataset\footnote{\url{https://github.com/waymo-research/waymo-open-dataset}.} \dotfill Apache License 2.0
    \item semantic-poss-api\footnote{\url{https://github.com/Theia-4869/semantic-poss-api}.} \dotfill MIT License
    \item SLidR\footnote{\url{https://github.com/valeoai/SLidR}.} \dotfill Apache License 2.0
    \item DINOv2\footnote{\url{https://github.com/facebookresearch/dinov2}.} \dotfill Apache License 2.0
    \item Segment-Any-Point-Cloud\footnote{\url{https://github.com/youquanl/Segment-Any-Point-Cloud}.} \dotfill CC BY-NC-SA 4.0
    \item torchsparse\footnote{\url{https://github.com/mit-han-lab/torchsparse}.} \dotfill MIT License
    \item ScaLR\footnote{\url{https://github.com/valeoai/ScaLR}.} \dotfill Apache License 2.0
    \item SuperFlow\footnote{\url{https://github.com/Xiangxu-0103/SuperFlow}} \dotfill Apache License 2.0
    \item FRNet\footnote{\url{https://github.com/Xiangxu-0103/FRNet}} \dotfill Apache License 2.0
\end{itemize}

\clearpage\clearpage
{
    \small
    \bibliographystyle{ieeenat_fullname}
    \bibliography{egbib}
}

\end{document}